\documentclass[twoside,11pt]{article}

\usepackage[bookmarks=true]{hyperref}
\hypersetup{
    colorlinks=false,       % false: boxed links; true: colored links
    linkcolor=blue,       % color of internal links
    citecolor=green,       % color of links to bibliography
    filecolor=red,        % color of file links
    urlcolor=blue,        % color of external links
    linktoc=page            % only page is linked
}
\usepackage{jair, theapa, rawfonts}

\usepackage{graphicx}
\usepackage{amsfonts}
\usepackage[boxed,ruled,vlined,linesnumbered,scleft]{algorithm2e}
\usepackage[dvipsnames]{xcolor}
\usepackage{amssymb}

\jairheading{x}{xxxx}{x-xx}{x/xx}{x/xx}
\ShortHeadings{Community Structure in Industrial SAT Instances}
{Ans\'otegui, Bonet, Gir\'aldez-Cru, Levy \& Simon}
\firstpageno{25}

\newcommand{\arity}{\mbox{deg}}

\newtheorem{theorem}{Theorem}

\newtheorem{definition}[theorem]{Definition}

\begin{document}

%\begin{frontmatter}

\title{Community Structure in Industrial SAT Instances}
 
\author{\name Carlos Ans\'otegui \email carlos@diei.udl.cat \\
       \addr DIEI, UdL, Jaume II 69,  25003 Lleida, Spain
       \AND
       \name Maria Luisa Bonet \email bonet@cs.upc.edu \\
       \addr CS, UPC, J.~Girona 1--3,  08034 Barcelona, Spain
       \AND
       \name Jes\'us Gir\'aldez-Cru \email jgiraldez@ugr.es \\
       \addr DaSCI, DECSAI, Universidad de Granada, 18071 Granada, Spain
       \AND
       \name Jordi Levy \email levy@iiia.csic.es \\
       \addr IIIA, CSIC, Campus UAB,  08193 Bellaterra, Spain
       \AND
       \name Laurent Simon \email lsimon@labri.fr \\
       \addr Univ. Bordeaux, CNRS, Bordeaux INP, LaBRI, UMR 5800, F-33400, Talence, France
}

%%%%%%%%%%%%%%%%%%%%%%%%%%%%%%%%%%%%%%%%%%%%%%%%%%%%%%%%%%%%%%%%%%%%%%%%%%%

\maketitle
  
\begin{abstract}
Modern SAT solvers have experienced a remarkable progress on solving industrial instances. Most of the techniques have been developed after an intensive experimental process.  It is believed that these techniques exploit the underlying structure of industrial instances. However, there are few works trying to exactly characterize the main features of this structure.

The research community on complex networks has developed techniques of analysis and algorithms to study real-world graphs that can be used by the SAT community. Recently, there have been some attempts to analyze the structure of industrial SAT instances in terms of complex networks, with the aim of explaining the success of SAT solving techniques, and possibly improving them.
  
In this paper, inspired by the results on complex networks, we study the \emph{community structure}, or \emph{modularity}, of industrial SAT instances.  In a graph with clear community structure, or high modularity, we can find a partition of its nodes into communities such that most edges connect variables of the same community. In our analysis, we represent SAT instances as graphs, and we show that most application benchmarks are characterized by a high modularity. On the contrary, random SAT instances are closer to the classical Erd\"os-R\'enyi random graph model, where no structure can be observed. We also analyze how this structure evolves by the effects of the execution of a CDCL SAT solver. In particular, we use the community structure to detect that new clauses learned by the solver during the search contribute to destroy the original  structure of the formula. This is, learned clauses tend to contain variables of distinct communities.

Motivated by this observation, we finally present an application that exploits the community structure to detect relevant learned clauses, and we show that detecting these clauses results in an improvement on the performance of the SAT solver. In particular, this application is presented as a preprocessing technique, so it can be easily incorporated into any existing solver. The preprocessor forces the solver to learn clauses having variables of two distinct communities, and keeps these clauses forever. Empirically, we observe that this improves the performance of several SAT solvers on industrial SAT formulas, especially on satisfiable instances. This suggests that the community structure of industrial SAT formulas is not a simple artifact, but it captures a relevant feature of the underlying structure of these instances, which partially explains the distinct performance of SAT solvers on random and industrial formulas, and which can be exploited by modern SAT solvers.
\end{abstract}

%\begin{keyword}
%  Satisfiability; modularity; Boolean formulas; complex networks.
%\end{keyword}

%\end{frontmatter}

%%%%%%%%%%%%%%%%%%%%%%%%%%%%%%%%%%%%%%%%%%%%%%%%%%%%%%%%%%%%%%%%%%%%%%%%%%%
\section{Introduction}\label{sec-intro}
%%%%%%%%%%%%%%%%%%%%%%%%%%%%%%%%%%%%%%%%%%%%%%%%%%%%%%%%%%%%%%%%%%%%%%%%%%%

The Boolean Satisfiability problem (SAT) is central in Computer
Science. Even though the general SAT problem is NP-Complete, many very
large industrial instances can be efficiently solved by modern SAT
solvers. Hence, SAT is extensively used to encode and solve many
problems from domains such as hardware and software verification, planning,
cryptography, scheduling, among others. Therefore, finding good
algorithms to solve SAT is of practical use in many areas of Computer
Science.

Although nowadays large real-world instances can be efficiently
solved, most relatively smaller random formulas cannot. It is
well-known in the SAT community that classical random $k$-CNF formulas
and industrial instances have a distinct nature. The intuition is that
the difference in performance of SAT solvers between random and
industrial instances comes from the existence of some kind of
\emph{structure} in industrial instances that can be
exploited~\cite{Hogg96,GomesS97,GentHPW99,WilliamsGS03,Jarvisalo2008,Laurent09}. 
As a result, SAT solvers tend to specialize in one or the other kind of formulas.
In fact, in the SAT competitions these formulas are evaluated in different
tracks. In the case of (almost) all application benchmarks,
Conflict-Driven Clause Learning (CDCL) SAT solvers show the best
performance, even when these instances come from very different
domains, as hardware verification, planning or cryptography. The main
component of these solvers is the learning of new clauses during the
search~\cite{KatebiSM11,SakallahM11}. The motivation of this work is to study the
body of industrial instances to detect general properties that are
shared by the majority of instances, even when they come from different domains. This knowledge can help
understand the success of CDCL SAT solvers on these benchmarks, and
possibly improve them.

The inspiration of our analysis comes from the works on \emph{complex
  networks} where the general structure of real-world graphs is
studied. To this effect, we use two ways to represent the SAT
instances as graphs. One model represents them as bipartite graphs,
where variables and clauses are nodes, and edges represent the
presence of a variable in a clause. In the second model, variables are
nodes, and edges between nodes (variables) indicate that there exists
a clause in which the two variables appear.

The classical \emph{Erd\"os-R\'enyi random graph
  model}~\cite{erdos-renyi} was one of the best studied during the
last century, and set the basis of graph theory. In this model, the
degree of nodes follows a binomial distribution. Random $k$-CNF
formulas, represented as graphs, follow this model. For instance, for
$k=3$, in the phase transition point, most of the variables have a
number of occurrences very close to $12.78$,\footnote{The number
  $12.78$ comes from multiplying the size of the clauses $k=3$ by the
  clause/variable ratio $m/n=4.26$ at the phase transition point.}
with a small variability in big graphs.  In the context of real-world
networks, other models have been defined.
  
A first model is the \emph{small-world} topology, proposed by
\citeA{strogatz}, as a new model to describe the structure of some
social networks. These networks are characterized by short path
lengths and high clustering factors.

Another is the \emph{scale-free} model, introduced by
\citeA{barabasi99} to describe the structure of the World Wide
Web. They show that the WWW, viewed as a graph, has a structure that
cannot be described by the classical random graph model. This means
that this graph is very different from what one would expect if edges
existed independently and at random. The name of this model comes from
the fact that, in this new model, the degree of nodes follows a
power-law distribution $P(k)\sim k^{-\alpha}$, and this distribution
is scale-free.

The topology of graphs has a major impact on the cost of solving
search problems on these graphs. \citeA{GentHPW99} analyze the impact
of a small-world topology on the cost of coloring graphs, and
\citeA{toby} does the same in the case of scale-free
graphs. \citeA{Walsh99} analyzes the small world topology of many
graphs associated with search problems in AI. He also shows that the
cost of solving these search problems can have a \emph{heavy-tailed
  distribution}. Therefore, we can expect that SAT solving, viewed as
a search process on a graph (the formula), will be affected by the
topology of this graph.

In this paper, we focus on the analysis of the community
structure. This is a very characteristic feature in real-world
networks~\cite{fortunato10}, that has received the attention of many
researchers in the last years. In order to analyze the community
structure of SAT instances, we use the notion of \emph{modularity}
introduced by~\citeA{newman04}. Having high modularity (in a graph)
means that nodes can be grouped into sets or communities, such that
there are many edges between nodes of the same community, but there
are few edges connecting nodes from different communities. Notice that the notion
of \emph{community} is more general than the notion of \emph{connected
  component}.  In particular, it allows the existence of (a few)
connections between communities. \citeA{BiereS06} show that many SAT
instances can be decomposed into \emph{connected components}, and how
to handle them within a SAT solver. They discuss how this structure into
\emph{connected components} can be used to improve the
performance of SAT solvers. Since our notion of community is more
general, it might be more practical to analyze and improve the
performance of SAT solvers.

The first contribution of this work is an exhaustive analysis of the
community structure of SAT instances. We show that industrial SAT
instances are characterized by a very clear community structure, i.e.,
high modularity. On the contrary, random formulas do not have
community structure, thus the modularity is very low (as
expected). Interestingly, this feature of SAT instances can be
computed with efficient algorithms. As we will see in the next
section, this result has been already used as the core of
other applications, as some modularity-based SAT and MaxSAT
solvers~\cite{DMartinsML13,SonobeKI14,NevesMJLM15}
or some modularity-based pseudo-industrial random
generators~\cite{IJCAI15,AIJ4,IJCAI17}.
These solvers have achieved some improvements by exploiting the community structure of the formula. The development of random generators capturing the properties of real-world SAT formulas is considered as a very challenging problem~\cite{tenchallenges1,tenchallenges2,tenchallenges3,tenchallenges4}. Using the previous random generators, which use the community structure, it has been observed that CDCL SAT solvers may exploit this community structure.

The second contribution is the analysis of the evolution of the
community structure during SAT solver search. In particular, we focus
on the effects of learning new clauses on this structure. We show that
learned clauses usually contain variables of distinct
communities. Therefore, the SAT solver \emph{tends} to destroy the
original partition of the formula. We remark that this result is very
interesting since it allows us to better understand the behavior of
the solver using a simple, compact feature: the community
structure. We consider that a better understanding of the success of
CDCL techniques is a required step to improve them.

Our last contribution is an application based on the previous observation.
We present it as a preprocessing technique, which exploits the community structure to detect relevant learned clauses. Detecting these clauses results in an improvement on the performance of the SAT solver.
In particular, the preprocessor first computes the community structure of the instance, i.e., it assigns each variable and each clause to a certain community. Then, it creates all subformulas composed by the clauses of two connected communities, and solves them, storing all learned clauses generated in this process. Notice that all these learned clauses only contain variables of (at most) two distinct communities. Notice also that, since all subformulas are extremely easy, this step is computed very quickly. Finally, it modifies the original formula by adding all these learned clauses, and the resulting formula is given as input to any actual SAT solver. We empirically evaluate this technique by running several SAT solvers with and without the preprocessor, and observe that in many cases their performance is improved when the preprocessing step is performed, especially on satisfiable instances.
Moreover, this contribution is also interesting because it shows the relation between the community structure and the relevance of learned clauses, and a way of exploiting it in practice.

Preliminary results of this paper have been presented in several conference publications~\cite{SAT12,SAT15}. Here, we extend our analysis including results on the community structure of a larger set of benchmarks: all industrial SAT instances that were used in the SAT Competitions from 2010 to 2017; and also an experimental evaluation of the modularity-based preprocessor on this larger set of formulas, including also new SAT solvers. The preliminary results of those conference publications are very similar to the new results observed in this more extensive analysis, reinforcing our conclusions.

The rest of the paper proceeds as follows. Related work and some
preliminary concepts are introduced in Sections~\ref{sec-related}
and~\ref{sec-preliminaries}, respectively. In
Section~\ref{sec-graphs}, we introduce the analysis of the community
structure in graphs, and our analysis of the community structure in
SAT instances is presented in Section~\ref{sec-community}. In
Section~\ref{sec-cdcl}, we show how this structure is affected by CDCL
techniques. Our preprocessing technique exploiting the community structure of the formula is presented in Section~\ref{sec-modprep}, where we include an extensive experimental evaluation of this preprocessor on several CDCL SAT solvers. Finally, conclusions are in Section~\ref{sec-conclusions}.

%%%%%%%%%%%%%%%%%%%%%%%%%%%%%%%%%%%%%%%%%%%%%%%%%%%%%%%%%%%%%%%%%%%%%%%%%%%
\section{Related Work}\label{sec-related}
%%%%%%%%%%%%%%%%%%%%%%%%%%%%%%%%%%%%%%%%%%%%%%%%%%%%%%%%%%%%%%%%%%%%%%%%%%%

The previous work on the community structure of SAT formulas~\cite{SAT12} has strongly influenced other works. The community structure is
correlated to the runtime of CDCL SAT
solvers~\cite{NewshamGFAS14,NewshamLGLFC15}. Also, it has been used to
improve the performance of several solvers. \citeA{DMartinsML13}
partition MaxSAT instances using the community structure in order to
identify smaller unsatisfiable subformulas. This method is refined
by~\citeA{NevesMJLM15}. \citeA{SonobeKI14} use the partition obtained
with the community structure to improve the performance of a parallel
SAT solver. The relation between community structure and BMC problems encoded into SAT instances has been studied~\cite{BaudBerthierGS17}.

An important issue to develop new SAT solving techniques
\emph{specialized} in industrial problems is the limited number of
these benchmarks and the high cost of solving them. For these reasons,
the generation of random instances with properties more similar to
industrial formulas is a very interesting challenge. This problem was
already stated by~\citeA{tenchallenges1} as one of the ten most
interesting challenges in propositional search. The same problem is
highlighted by~\citeA{tenchallenges3} and~\citeA{dechterbook}. Some approaches on
pseudo-industrial random generation focus on general properties
shared by the majority of real-world problems. This is the case of the
(clear) community structure. There exist some generators that
\emph{indirectly} use the notion of
modularity~\cite{Slater02,BurgKaufmannKottler2012,NewshamGFAS14,MalitskyMOT16}. Recently,
the Community Attachment model~\cite{IJCAI15,AIJ4} has been proposed
to generate random pseudo-industrial instances with high modularity.
\citeA{ijcai09,AIJ5} have proposed a model for generating scale-free random SAT instances. \citeA{IJCAI17} extended it to the Popularity-Similarity model for SAT instances, a model where the notion of modularity (similarity) can be combined with the high variability of variable occurrences (popularity).

The underlying structure of SAT instances and its relations to the
performance of SAT solvers have been also addressed in other related
works.  Most industrial SAT instances have a scale-free
structure~\cite{CP09}. In particular, it is shown that the number of
variable occurrences $k$ follows a power-law distribution $P(k) \sim
k^{-\alpha}$. \citeA{pagerank} study the centrality of branching variables
selected by a CDCL solver. \citeA{Simon14} uses observations from the
SAT solver performance on industrial problems to better understand its
behavior. Also, most industrial SAT instances have fractal
dimension~\cite{IJCAR14}. This means that the shape of the graph is
preserved after rescaling, i.e., replacing groups of nodes by a single
node.

%%%%%%%%%%%%%%%%%%%%%%%%%%%%%%%%%%%%%%%%%%%%%%%%%%%%%%%%%%%%%%%%%%%%%%%%%%%
\section{Preliminaries}\label{sec-preliminaries}
%%%%%%%%%%%%%%%%%%%%%%%%%%%%%%%%%%%%%%%%%%%%%%%%%%%%%%%%%%%%%%%%%%%%%%%%%%%

Given a set of Boolean variables $X=\{x_1, \ldots, x_n\}$, a
\emph{literal} is an expression of the form $x_i$ or $\neg x_i$.  A
\emph{clause} $c$ of size $s$ is a disjunction of $s$ literals,
$l_1 \vee \ldots \vee l_s$.  We note $s = |c|$, and say that $x\in c$,
if $c$ contains the literal $x$ or $\neg x$.  A \emph{CNF formula}
or \emph{SAT instance} of length $t$ is a conjunction of $t$
clauses, $c_1 \wedge \ldots \wedge c_t$.  A \emph{k-CNF formula} is
a conjunction of $k$-sized clauses.

An (undirected) weighted graph is a pair $(V,w)$ where $V$ is a set of
vertexes and $w:V\times V\to\mathbb{R}^+$ satisfies $w(x,y)=w(y,x)$.
This definition generalizes the classical notion of graph $(V,E)$,
where $E\subseteq V\times V$, by taking $w(x,y) = 1$ if $(x,y)\in E$
and $w(x,y)=0$ otherwise.  The degree of a vertex $x$ is defined as
$\arity(x) = \sum_{y\in V}w(x,y)$.  A bipartite graph is a tuple
$(V_1,V_2,w)$ where $V_1$ and $V_2$ are two disjoint sets of vertexes,
and $w: V_1\times V_2\to\mathbb{R}^+$.

Given a SAT instance, we construct two graphs, following two models.
In the Variable Incidence Graph model (VIG, for short), vertexes
represent variables, and edges represent the existence of a clause
relating two variables. A clause $l_1\vee\dots\vee l_n$ results into
$n\choose 2$ edges, one for every pair of variables in the clause. Notice also that
there can be more than one clause relating two given variables. To
preserve this information we put a higher weight on edges connecting
variables related by more clauses.  Moreover, to give the same
relevance to all clauses, we ponder the contribution of a clause to
an edge by $1/{n\choose 2}$. This way, the sum of the weights of the
edges generated by a clause is always one.

\begin{definition}[Variable Incidence Graph (VIG)]
  Given a SAT instance $\Gamma$ over the set of variables $X$, its
  variable incidence graph is a graph $(X,w)$ with set of vertexes the
  set of Boolean variables, and weight function:
$$
w(x,y) = \sum_{c\in \Gamma \atop x,y\in c} 
\frac{1}
{{|c| \choose 2}}
$$
\label{def-vig}
\end{definition}

In the Clause-Variable Incidence Graph model (CVIG, for short),
vertexes represent either variables or clauses, and edges represent
the occurrence of a variable in a clause. Like in the VIG model, we
try to give the same relevance to all clauses, thus every edge
connecting a variable $x$ with a clause $c$ containing it has weight
$1/|c|$. This way, the sum of the weights of the edges generated by a
clause is also one in this model.

\begin{definition}[Clause-Variable Incidence Graph (CVIG)]
  Given a SAT instance $\Gamma$ over the set of variables $X$, its
  clause-variable incidence graph is a bipartite graph $(X,\{c\mid
  c\in\Gamma\},w)$, with vertexes the set of variables and the set of
  clauses, and weight function:
$$
w(x,c) = \left\{
  \begin{array}{ll}
    1/|c|&\mbox{if $x\in c$}\\
    0&\mbox{otherwise}
  \end{array}\right.
$$
\label{def-cvig}
\end{definition}

From now on we will indistinctly use the words formula or graph to discuss
SAT formulas.

%%%%%%%%%%%%%%%%%%%%%%%%%%%%%%%%%%%%%%%%%%%%%%%%%%%%%%%%%%%%%%%%%%%%%%%%%%%
\section{The Community Structure of Graphs}\label{sec-graphs}
%%%%%%%%%%%%%%%%%%%%%%%%%%%%%%%%%%%%%%%%%%%%%%%%%%%%%%%%%%%%%%%%%%%%%%%%%%%

The notion of \emph{modularity} was introduced
by~\citeA{newmangirvan04}.  This property is defined for a graph and a
specific \emph{partition} of its vertexes into \emph{communities}, and
measures the density of internal edges, i.e., edges between nodes of
the same community.  Thus, in a graph with high modularity, there
exists a partition of its nodes such that most of the edges connect
nodes of the same community.  The modularity of a graph is then the
maximal modularity for all possible partitions of its vertexes.
Obviously, measured this way, the maximal modularity would be obtained
putting all vertexes in the same community. To avoid this problem,
\citeA{newmangirvan04} define modularity as the fraction of edges
connecting vertexes of the same community minus the \emph{expected}
fraction of edges in a random graph with the same number of vertexes
and the same node degrees.

\begin{definition}[Modularity of a Graph]  
  Given a weighted graph $G=(V,w)$ and a partition $P=\{P_1,\dots,P_n\}$ of
  its vertexes $V$, we define their \emph{modularity} as
$$
Q(G,P) = \sum_{P_i\in P} \frac{\displaystyle\sum_{x,y\in P_i} w(x,y)}
{\displaystyle\sum_{x,y\in V} w(x,y)} -
\left(
\frac{\displaystyle\sum_{x\in P_i}\arity(x)}
{\displaystyle\sum_{x\in V}\arity(x)}
\right)^2
$$

The \emph{(optimal) modularity} of a graph is the maximum modularity,
for any possible partition of its vertexes: $Q(G)=\max\{Q(G,P)\mid P\}$
\end{definition}

Since both terms in the definition of modularity are in the range $[0,
  1]$, and, for the partition given by a single community, both have
value $1$, the optimal modularity of graph will be in the range $[0,
  1]$. In practice, $Q$ values for networks showing a strong community
structure range from 0.3 to 0.7, higher values are
rare~\cite{newmangirvan04}.

There has not been an agreement on the definition of modularity for
bipartite graphs. Here we will use the notion proposed
by~\citeA{barber07} that extends Newman and Girvan's definition by
restricting the random graphs used in the second term of such
definition to be bipartite. In this new definition, communities may
contain vertexes of both sets $V_1$ and $V_2$.

\begin{definition}[Modularity of a Bipartite Graph]  
  Given a graph $G=(V_1,V_2,w)$ and a partition $P=\{P_1,\dots,P_n\}$ of
  its vertexes $V_1\cup V_2$, we define their \emph{modularity} as
$$
Q(G,P) = \sum_{P_i\in P} \frac{\displaystyle\sum_{x\in P_i\cap V_1\atop y\in P_i\cap V_2} w(x,y)}
{\displaystyle\sum_{x\in V_1\atop y\in V_2} w(x,y)}\ \ -\ \
\frac{\displaystyle\sum_{x\in P_i\cap V_1}\arity(x)}
{\displaystyle\sum_{x\in V_1}\arity(x)}\cdot
\frac{\displaystyle\sum_{y\in P_i\cap V_2}\arity(y)}
{\displaystyle\sum_{y\in V_2}\arity(y)}
$$
\label{def-modbi}
\end{definition}

There exist a wide variety of algorithms for computing the modularity
of a graph. Moreover, there exist alternative notions and definitions
of modularity for analyzing the community structure of a network.~\citeA{fortunato10} presents a survey in the field. The decision version of
modularity maximization is NP-complete~\cite{brandes08}.  Therefore,
all efficient modularity-optimization algorithms proposed in the
literature, instead of computing the exact value of the modularity,
return an approximation of $Q$, in fact a lower bound of $Q$. They
include greedy methods, methods based on simulated annealing, on
spectral analysis of graphs, etc.  Most of them have a complexity that
make them inadequate to study the structure of very large graphs, like
industrial SAT instances. There are algorithms especially designed to
deal with large-scale networks, like the greedy algorithms for
modularity optimization~\cite{newman04,clausetetal04}, the label
propagation-based algorithm~\cite{Raghavan} and the method based on
graph folding~\cite{fastmodularity}.

\SetKwBlock{function}{}{}
\begin{algorithm}[t]

\KwIn{Graph $G=(X,w)$}
\KwOut{Label $L_1$}
\ForEach{$i\in X$}{$L_1[i] := i$}
$L_2 := Labelling(G)$\;
\While{Modularity$(G,L_1)<$ Modularity$(G,L_2)$}{
  $L_1 := L_2$\tcp*{labelling of original nodes in the folded graph}
  $G := Folding(G,L_2)$\;
  $L_2 := Labelling(G)$\;
}  

{\bf function} $Labelling(Graph\ G=(X,w)): Label\ L$
\function {
  \ForEach{$i\in X$}{$L[i] := i$}
  \Repeat{$\neg changes$}{
    $changes := false$\;
    \ForEach{$i \in X$}{
      $bestinc :=0$\;
      \ForEach{$c \in  \{c\mid \exists j. w(i,j)\neq 0 \wedge L[j] = c\}$}{
        $inc := \sum_{L(j)=c}w(i,j) - deg(i) \cdot \sum_{L[j]=c} deg(j) / \sum_{j\in X} deg(j)$\;
        \If{$inc>bestinc$}{$L[i] := c$; $bestinc:=inc$; $changes := true$;}
      }
    }
  }
  {\bf return $L$}
}
{\bf function} $Folding(Graph\ G_1=(X,w), Label\ L): Graph\ G'$
\function {
  $X'=\{c\subseteq X\mid \forall i,j\in c\ .\  L[i]=L[j] \wedge c\mbox{ is maximal}\}$\;
  $w'(c_1,c_2) = \sum_{i\in c_1, j\in c_2} w(i,j)$\;
  {\bf return} $G'=(X',w')$;
}
\caption{Louvain Method (LM)~\cite{fastmodularity}}\label{alg-GFA}
\end{algorithm}

\begin{algorithm}[ht]
{\bf function} $Folding(Graph\ G_1=(X_1,X_2,w), Label\ L): Graph\ G'$
\function {
  $X_1'=\{c\subseteq X_1\mid \forall i,j\in c\ .\  L[i]=L[j]\wedge c\mbox{ is maximal}\}$\;
  $X_2'=\{c\subseteq X_2\mid \forall i,j\in c\ .\  L[i]=L[j]\wedge c\mbox{ is maximal}\}$\;
  $w'(c_1,c_2) = \sum_{i\in c_1, j\in c_2} w(i,j)$\;
  {\bf return} $G'=(X_1',X_2',w')$;
}
\caption{Folding function for bipartite graphs}\label{alg-GFA2}
\end{algorithm}

The first algorithm for modularity maximization was described
by~\citeA{newman04}. This algorithm starts by assigning every vertex
to a distinct community. Then, it proceeds by joining the pair of
communities that results in a bigger increase of the modularity
value. The algorithm finishes when no community joining results in an
increase of the modularity. In other words, it is a greedy
gradient-guided optimization algorithm. The algorithm may also return
a dendrogram of the successive partitions found. Obviously, the
obtained partition may be a local maximum. \citeA{clausetetal04}
optimize the data structures used in this basic algorithm, using among
others, data structures for sparse matrices. The complexity of this
refined algorithm is $\mathcal{O}(m\,d\,\log n)$, where $d$ is the
depth of the dendrogram (i.e. the number of joining steps), $m$ the
number of edges and $n$ the number of vertexes.  They argue that $d$
may be approximated by $\log n$, assuming that the dendrogram is a
balanced tree, and the sizes of the communities are similar. However,
this is not true for the graphs we have analyzed, where the sizes of
the communities are not homogeneous. This algorithm has not been able
to finish, for any of our SAT instances, with a runtime limit of one
hour.

An alternative algorithm is the \emph{Label Propagation Algorithm
  (LPA)} proposed by~\citeA{Raghavan}. Initially, all vertexes are
assigned to a distinct label, e.g., its identifier. Then, the
algorithm proceeds by re-assigning to every vertex the most
frequent label among its neighbors. The procedure ends when every
vertex is assigned a label that is maximal among its neighbors.  In
case of a tie between the most frequent labels, the winning label is
chosen randomly.  The algorithm returns the partition defined by the
vertexes sharing the same label. The label propagation algorithm has a
near linear complexity.  However, it has been shown experimentally
that the partitions it computes have a worse modularity than the
partitions computed by the Newman's greedy algorithm.

The \emph{Louvain Method (LM)}\footnote{In some works, this method is
  also known as Graph Folding Algorithm (GFA).} proposed
by~\citeA{fastmodularity} (see Alg.~\ref{alg-GFA}) improves the Label
Propagation Algorithm in two directions. The idea of moving one node
from one community to another following a greedy strategy is the same,
but instead of selecting the community where the node has more
neighbors, it selects the community where the movement would most
increase the modularity. Second, once no movement of nodes from
community to community can increase the modularity (we have reached a
local modularity maximum), we allow to merge communities, in order to repeat this process multiple
times. In more details, the algorithm proceed as follows. Initially, we have a labelling (baseline) with every node assigned to a distinct community (lines 1-2). Then, two steps are repeated. First, the labelling procedure is executed (line 3, 8-20). In this process, we start again with the initial labelling (lines 9-10). It then randomly iterates among all the nodes of the graph (lines 11-13), checking whether moving a node to the community of any of its neighbors (line 15) produces a modularity higher than the current one (line 16), and updating the labelling accordingly (lines 17-18). These iterations among all the nodes of the graph are repeated until no node changes its labelling. The new labelling replaces the initial one when the modularity is increased (lines 4-5). Second, the existing graph is folded (lines 6, 21-24). In this process, we construct a new graph
where we have one node for each community of the original graph (line~22), and we include an edge between nodes $c_1$ and $c_2$ if in the original graph there exists an edge between a node in the community $c_1$ and a node in the community $c_2$. In the case of weighted graphs, the weight of the edge $w'(c_1,c_2)$ in the new graph is $\sum_{i\in c_1,j\in c_2} w(i,j)$, where $w(i,j)$ is the weight in the original graph (line~23). Notice that in Alg.~\ref{alg-GFA}, line 22, the set $X'$ is the partition in $X$ induced by the set of labels. Finally, we repeat these two steps until the value of the modularity is not improved any more.

Notice, however, that LM is not able
to compute the community structure of bipartite graphs according to
Definition~\ref{def-modbi}. This is because after the first folding, LM would collapse all nodes of the same
community into a single node in the folding step, destroying the
bipartite structure of the graph. Therefore, in order to compute the
community structure of the CVIG model, we have adapted this algorithm
to bipartite graphs, re-implementing the folding step to preserve the
bipartite structure of the graph. In particular, we replace the
folding function by the function described in
Algorithm~\ref{alg-GFA2}.
This function differs from the folding function in Alg.~\ref{alg-GFA} in the nodes that are collapsed into a new node. In particular, in order to preserve the bipartite structure of the graph we distinguish two sets $X'_1$ and $X'_2$ (lines 2-3), one for each partition of the bipartite graph. Again, these sets $X'_1$ and $X'_2$ are respectively the partitions in $X_1$ and $X_2$ induced by the set of labels.

In our experiment,
we use this method since it gives better bounds in both models VIG and
CVIG than other algorithms, like LPA~\cite{SAT12}.

%%%%%%%%%%%%%%%%%%%%%%%%%%%%%%%%%%%%%%%%%%%%%%%%%%%%%%%%%%%%%%%%%%%%%%%%%%%
\section{The Community Structure of Industrial SAT Instances}\label{sec-community}
%%%%%%%%%%%%%%%%%%%%%%%%%%%%%%%%%%%%%%%%%%%%%%%%%%%%%%%%%%%%%%%%%%%%%%%%%%%

In this section, we present the analysis of the community structure of
SAT instances. To this purpose, we represent SAT instances as graphs
using the VIG and CVIG model, and we analyze the community structure
of these graphs using the Louvain Method.\footnote{The software we use in the
experimentation is publicly available in {\small
  \url{https://www.ugr.es/~jgiraldez/}}.}

\begin{table}[t]
\begin{center}
\begin{tabular}{|rr||r|r|r|r|}
\hline
\multicolumn{1}{|c}{$n$} & \multicolumn{1}{c||}{$m/n$} & \multicolumn{1}{c}{$Q$} & \multicolumn{1}{|c}{$|P|$} & \multicolumn{1}{|c}{$larg$} & \multicolumn{1}{|c|}{$iter$} \\
\hline
\hline
10$^4$ & 1.00 & \bf 0.486 & 545 & 3.8 & 54 \\
10$^4$ & 1.50 & 0.353 & 146 & 5.1 & 52 \\
10$^4$ & 2.00 & 0.280 & 53 & 6.8 & 51 \\
10$^4$ & 3.00 & 0.217 & 14 & 15.5 & 64 \\
10$^4$ & 4.00 & 0.178 & 11 & 14.8 & 54 \\
10$^4$ & 4.25 & 0.170 & 11 & 14.6 & 53 \\
10$^4$ & 4.50 & 0.163 & 11 & 14.7 & 53 \\
10$^4$ & 5.00 & 0.152 & 11 & 14.3 & 51 \\
10$^4$ & 6.00 & 0.133 & 12 & 13.9 & 53 \\
10$^4$ & 7.00 & 0.120 & 10 & 15.0 & 56\\
10$^4$ & 8.00 & 0.138 & 6 & 25.0 & 50\\
10$^4$ & 9.00 & 0.130  & 6 & 24.3 & 49\\
10$^4$ & 10.00 & 0.123 & 6 & 24.4 & 47\\
\hline
\end{tabular}
\end{center}
\caption{Modularity of random 3-CNF formulas varying the
  clause/variable ratio $m/n$, for $n=10^4$
  variables. Results are computed for the LM algorithm on the VIG model.}\label{tab-static-community-random-1}
\end{table}

\begin{table}[t]
\begin{center}
\begin{tabular}{|rr||r|r|r|r|}
\hline
\multicolumn{1}{|c}{$n$} & \multicolumn{1}{c||}{$m/n$} & \multicolumn{1}{c}{$Q$} & \multicolumn{1}{|c}{$|P|$} & \multicolumn{1}{|c}{$larg$} & \multicolumn{1}{|c|}{$iter$} \\
\hline
\hline
10$^2$ & 4.25 & 0.170 &   6.3 & 23.6 &   10 \\
10$^3$ & 4.25 & 0.181 & 10.4 & 17.4 &   38 \\
10$^4$ & 4.25 & 0.166 & 11.2 & 15.0 &   54 \\
10$^5$ & 4.25 & 0.148 & 19.1 & 11.8 & 102 \\
10$^6$ & 4.25 & 0.145 & 29.6 & 13.0 & 171 \\
\hline
\end{tabular}
\end{center}
\caption{Modularity of random 3-CNF formulas at the peak
  transition region (clause/variable ratio $m/n$=4.25), varying the
  number of variables $n$. Results computed for the LM algorithm on the VIG model.}\label{tab-static-community-random-2}
\end{table}

First, we analyze the community structure of random SAT instances. We recall that random formulas (or random graphs) are not expected to exhibit any structure at all. However, we can use these results as a baseline to \emph{measure} how clear is the community structure in industrial SAT instances. Since the baseline must be as good as possible, we only report the results of our analysis on the VIG model because, as we will show later, the performance of the LM algorithm is worse in bipartite graphs (i.e., in the CVIG model).

In this analysis, we report the modularity $Q$ of the partition
returned by the Louvain Method, as well as the number of communities
$|P|$ and the percentage $larg$ of nodes belonging to the largest
community. We also report the number of iterations $iter$ spent by
the LM algorithm, being each iteration an execution of the main loop
of the function $Labelling$. Notice that each iteration visits all
nodes of the graph. Therefore, this number gives an intuition about
the runtime of the LM on SAT instances.

In the first experiment on random formulas, we study the modularity of random 3-CNF SAT
instances with different clause to variable ratio $m/n$, for a fixed
number of variables $n=10^4$. For every value of $m/n$, we generate $100$ instances, and we report
the average results (as expected, there is no significant dispersion on these results).
Table~\ref{tab-static-community-random-1} shows the results. As
we can see, the modularity of random instances is only significant for
very low clause/variable ratios, i.e., on the leftist satisfiable easy
side. This is due to the presence of a large quantity of very small
unconnected components, which make a clear division into communities of the nodes (variables) of the graph (formula).
Even though, for these low values of $m/n$,
the modularity is not as high as for industrial instances, as we will
see later, confirming their distinct nature. Notice that as the
clause/variable ratio $m/n$ increases, the variables get more
connected but without following any particular structure, and the
number of communities highly decreases. This explains the low value of
the modularity for this family of benchmarks. Also, we do not observe
any abrupt change in the phase transition point.

As a second experiment with random 3-SAT instances, we want to
investigate the modularity of these formulas at the peak transition region $m/n=4.25$,
for an increasing number of variables
$n$. Table~\ref{tab-static-community-random-2} shows the results. As
we can see, the modularity is very low and it tends to slightly
decrease as the number of variables increases, and seems to tend to a
particular value ($0.15$ for the phase transition point).

We observe that in almost all the cases, the modularity $Q$ of random 3-SAT formulas is lower than $0.2$.
From now on, we will mark in bold the values of $Q$ greater than the double of this value, i.e., when $Q>0.4$.
We recall that the optimal value of the modularity of any graph is always in the interval $\left [0,1 \right ]$. 
Moreover, real-world networks exhibiting a clear community structure usually have a modularity in the interval $\left [0.3,0.7 \right ]$~\cite{newmangirvan04}.
Therefore, we consider our choice of $0.4$ is a reasonable bound, although any other reasonable value could have been selected instead.

Now, we present our analysis of the community structure of industrial SAT instances. We divide it in two parts.
First, we present detailed results using the set of industrial formulas of the SAT Competition
2013\footnote{\url{http://satcompetition.org/2013/}}.
This study allows us to analyze specific questions in more details.
Second, we report aggregated results using all industrial SAT instances that were 
used in any SAT Competition between 2010 and 2017.
This way, we verify that our previous results on the industrial instances of the SAT Competition 2013 are not a consequence of the selected benchmarks, but a general feature of industrial SAT formulas.

\begin{table}[t]
  \begin{center}
\tabcolsep 1mm
\begin{tabular}{|l@{\hspace{-1ex}}r||r|r|r|r|r||r|r|r|r|r||r|r|}\hline
&&
\multicolumn{5}{c||}{VIG}&
\multicolumn{5}{c||}{CVIG}&
\multicolumn{2}{c|}{CC} \\
Family & \kern -2mm\#inst. & $Q_{orig}$ & $Q_{prep}$ & \multicolumn{1}{c|}{$|P|$} & \multicolumn{1}{c|}{$larg$} & \multicolumn{1}{c||}{$iter$}&
$Q_{orig}$ & $Q_{prep}$ & \multicolumn{1}{c|}{$|P|$} & \multicolumn{1}{c|}{$larg$} & \multicolumn{1}{c||}{$iter$} &
\multicolumn{1}{c|}{$|P|$} & \multicolumn{1}{c|}{$larg$} \\\hline\hline
2d-strip-packing 	& 5 & \bf 0.942 & \bf 0.942 & 40.2 & 4.83 & 6.4 & \bf 0.932 & \bf 0.928 & 9835.0 & 3.36 & 8.6 & 1.0 & 100.0 \\
bio 				& 5 & \bf 0.607 & \bf 0.549 & 42.4 & 7.94 & 15.2 & 0.370 & 0.361 & 5994.8 & 0.20 & 7.6 & 1.4 & 99.9 \\
crypto-aes 		& 11 &\bf  0.804 & \bf 0.752 & 23.3 & 12.71 & 23.9 & \bf 0.610 & \bf 0.563 & 7379.3 & 4.05 & 18.5 & 1.0 & 100.0 \\
crypto-des 		& 9 & \bf 0.952 & \bf 0.929 & 82.4 & 2.94 & 19.8 &  \bf 0.498 &  \bf 0.473 & $>10^4$ & 0.03 & 12.2 & 1.0 & 100.0 \\
crypto-gos 		& 30 & \bf 0.639 & \bf 0.641 & 39.6 & 16.32 & 15.7 & \bf 0.633 & \bf 0.623 & 506.2 & 10.45 & 12.1 & 1.0 & 100.0 \\
crypto-md5 		& 11 & \bf 0.784 & \bf 0.780 & 33.1 & 6.06 & 40.5 & \bf 0.510 & \bf 0.544 & $>10^4$ & 0.03 & 16.6 & 1.0 & 100.0 \\
crypto-sha 		& 30 & \bf 0.558 & \bf 0.641 & 13.7 & 11.61 & 25.7 & \bf 0.562 & \bf 0.584 & 1001.5 & 0.20 & 10.7 & 1.0 & 100.0 \\
crypto-vmpc 		& 8 & 0.239 & 0.239 & 9.5 & 16.03 & 9.6 & 0.398 & 0.398 & 1047.3 & 0.25 & 6.8 & 1.0 & 100.0 \\
diagnosis 			& 26 & \bf 0.932 & \bf 0.927 & 56.8 & 4.45 & 42.3 &  \bf 0.483 &  \bf 0.444 & $>10^5$ & 0.01 & 18.5 & 1.0 & 100.0 \\
hardware-bmc-ibm\hspace{-5mm} 	& 4 & \bf 0.971 & \bf 0.956 & 76.0 & 2.52 & 37.5 &  \bf 0.499 &  \bf 0.468 & $>10^5$ & 0.03 & 33.5 & 1.0 & 100.0 \\
hardware-bmc 		& 3 & \bf 0.922 & \bf 0.886 & 20.7 & 7.65 & 29.3 &  \bf 0.496 &  \bf 0.432 & $>10^4$ & 0.07 & 18.0 & 1.0 & 100.0 \\
hardware-cec 		& 30 & \bf 0.857 & \bf 0.785 & 29.2 & 14.94 & 106.3 &  \bf 0.478 &  \bf 0.461 & $>10^4$& 1.06 & 85.9 & 1.1 & 99.9 \\
hardware-velev 	& 21 & \bf 0.679 & \bf 0.678 & 16.4 & 36.31 & 25.7 &  \bf 0.486 &  \bf 0.488 & $>10^5$ & 2.92 & 31.8 & 1.0 & 100.0 \\
planning 			& 25 & \bf 0.865 & \bf 0.850 & 22.6 & 9.85 & 24.2 &  \bf 0.497 &  \bf 0.496 & $>10^5$ & 0.01 & 41.6 & 1.0 & 100.0 \\
scheduling-pesp 	& 30 & \bf 0.780 &\bf  0.781 & 14.7 & 17.03 & 58.6 &  0.359 & 0.359 & $>10^4$ & 0.04 & 17.8 & 2.4 & 95.3 \\
scheduling 		& 30 & \bf 0.894 & \bf 0.892 & 45.7 & 6.12 & 178.7 &  \bf 0.474 &  \bf 0.456 & $>10^5$ & 0.01 & 66.8 & 1.0 & 100.0 \\
software-bit-verif 	& 12 & \bf 0.878 & \bf 0.801 & 21.0 & 9.85 & 45.3 & \bf 0.506 & \bf 0.568 & $>10^4$ & 2.49 & 57.4 & 1.0 & 100.0 \\
termination 		& 5 & \bf 0.775 & \bf 0.695 & 38.4 & 13.95 & 30.2 & \bf 0.525 & \bf 0.525 & $>10^4$ & 1.03 & 36.0 & 1.0 & 100.0 \\
\hline
\end{tabular}
\end{center}
\caption{Modularity before and after preprocessing, $Q_{orig}$ and $Q_{prep}$ respectively, for both VIG and CVIG of the industrial
  families of the SAT Competition 2013. We also include the analysis of the connected components (CC). $|P|$ stands for number of
  communities (or connected components), $larg$ for fraction of
  vertexes in the largest community (component), and $iter$ for number
  of iterations of the algorithm LM.}
\label{tab-static-community-families}
\end{table}

In Table~\ref{tab-static-community-families}, we report results of the
community structure of industrial SAT instances of the SAT Competition 2013, grouped by
families. For each family of industrial instances, we present the
results of the modularity $Q_{orig}$ of the original formulas, and the
modularity $Q_{prep}$ of these formulas after preprocessing with
Satelite~\cite{satelite} with default options.
The results about the number of communities ($|P|$), the percentage of vertexes belonging to the largest community ($larg$), and number of iterations of the algorithm
($iter$) correspond to the results with the preprocessed instances.
Finally, we also study the connected components, as suggested by~\citeA{BiereS06}.
We performed these experiments with a limit of 64GB RAM, obtaining results for all instances except the 3 formulas of the family \emph{software-bmc}, which are extremely huge. We omit this family in the table.
 
 The reason to group the results by families is that we observed that all instances of the same family have a similar
community structure (modularity, number of communities, etc..). For
instance, the maximal dispersion of the modularity $Q$ is found in the
family \emph{hardawre-velev} for the VIG model, with a standard
deviation $SD[Q]=0.0081$, which is a extremely low value.

We have to remark that the LM algorithm returns a lower bound on the
modularity. Having this in mind, we can conclude that, except for the
\emph{crypto-vmpc} family, all families show a very clear community
structure with values of $Q$ around $0.8$.  In other kind of networks,
values greater than $0.7$ are rare, therefore the values obtained for
industrial SAT instances can be considered as exceptionally high.

If we compare the modularity for the VIG model with the same values
for the CVIG model, we can conclude that, in general, these values are
higher for the VIG model. This is an effect of the LM algorithm when
it is applied to bipartite graphs.
In particular, in bipartite graphs, the folding procedure does not collapse every node of a community into a new node, but into two, with an edge of high weight between these two nodes. The existence of this edge makes that after the first folding, almost no node changes its labelling (in the next $Labelling$ procedure). As a consequence, LM stops.
Since the algorithm stops \emph{earlier} for the CVIG model, the number of iterations $iter$ is smaller and the number of communities $|P|$ is bigger.

We also compare the values of the modularity before and after
preprocessing the instances, $Q_{orig}$ and $Q_{prep}$
respectively. We see that in most cases, $Q_{prep}$ is slightly
smaller than $Q_{orig}$, and in some \emph{crypto} families, it is
even bigger. However, both values are very close. Therefore, we can
conclude that the default preprocessing techniques applied by Satelite
almost do not affect the community structure of the formula.

If all communities have a similar size, then $larg \approx 1/|P|$.  In
many cases in Table~\ref{tab-static-community-families}, we have $|P|
\gg 1/larg$. This means that the community structure has a big
variability in the sizes of the communities obtained.

With respect to the number of iterations, with the LM algorithm, in every
iteration we have to visit all neighbors of every node. Therefore, the
cost of an iteration is linear in the number of edges of the
graph. Moreover, after folding the graph, we can do further
iterations, and even several graph foldings.

Finally, we have also studied the \emph{connected components} of these
instances after preprocessing. As we can see in
Table~\ref{tab-static-community-families}, almost all instances have a
single connected component, i.e., almost all variables are included in
the same connected component. Hence the rest of connected components
contain just an insignificant subset of the variables. Therefore, the
modularity gives us much more information about the structure of the
formula than connected components. Notice that a connected component
can be structured into several communities. We also found a large
number of very small connected components in some industrial formulas
before preprocessing (these results are not shown in
Table~\ref{tab-static-community-families}). However, these components
are easily removed by the preprocessor.

We have observed that most of the industrial SAT instances in the set of
benchmarks previously analyzed is characterized by a clear community structure.
We recall, again, that the LM returns a lower-bound on the modularity.
Even if it is not the optimal value, when it is high enough we can use it to
guarantee a clear community structure.

The natural question now is whether this observation (i.e., these industrial SAT instances
are characterized by a clear community structure) is extensible to most of the application
SAT benchmarks\footnote{We restrict our analysis to the application benchmarks commonly used in the
SAT community, i.e., the benchmarks of the SAT Competitions.}. To this purpose, we carry out an analysis of
all applications benchmarks that have been used in the SAT Competitions between 2010 and 2017,
both included. This set contains a total of 2550 industrial SAT instances. In this experiment, we limit the RAM memory usage to 16GB.

\begin{table}[t]
\begin{center}
\begin{tabular}{|lr|r|r|r|r|r|r|}
\hline
Benchmark & \#inst. & $avg$ & $std$ & $med$ & $min$ & $p_{10\%}$ & MO \\
\hline
SAT Race 2010 			& 100 & \bf 0.829 & 0.15 & \bf 0.891 & 0.237 & \bf 0.671 & 3.0\% \\ %MO=3
SAT Competition 2011 	& 300 & \bf 0.836 & 0.13 & \bf 0.877 & 0.222 & \bf 0.689 & 2.0\% \\ %MO=6
SAT Challenge 2012 		& 600 & \bf 0.835 & 0.14 & \bf 0.885 & 0.231 & \bf 0.633 & 1.8\% \\ %MO=11
SAT Competition 2013 	& 300 & \bf 0.775 & 0.16 & \bf 0.813 & 0.231 & \bf 0.550 & 2.7\% \\ %MO=8
SAT Competition 2014 	& 300 & \bf 0.772 & 0.24 & \bf 0.837 & 0.050 & \bf 0.544 & 2.0\% \\ %MO=6
SAT Race 2015 			& 300 & \bf 0.746 & 0.19 & \bf 0.794 & 0.061 & \bf 0.457 & 8.0\% \\ %MO=24
SAT Competition 2016 	& 300 & \bf 0.818 & 0.15 & \bf 0.857 & 0.176 & \bf 0.582 & 7.4\% \\ %MO=26
SAT Competition 2017 	& 350 & \bf 0.668 & 0.26 & \bf 0.688 & 0.182 &     0.197 & 0.0\% \\ %MO=0
\hline
SAT Compts.	 2010-2017	& 2550 & \bf 0.785 & 0.19 & \bf 0.846 & 0.049 & \bf 0.545 & 3.7\% \\ %MO=84
\hline
\end{tabular}
\end{center}
\caption{Statistics about the modularity $Q$ of the VIG of all industrial SAT instances used in the SAT Competitions from 2010 to 2017. MO stands for memory-out.}
\label{tab-modularity-aggregated}
\end{table}

In Table~\ref{tab-modularity-aggregated}, we report some statistics (average, standard deviation, median, minimum and percentile 10\%) about the modularity $Q$ of the VIG of this set of benchmarks. We also report in this table the number of instances for which we could not compute the community structure due to the large amount of memory required for this computation.

It can be observed that both the average and the median of the modularity $Q$ is extraordinarily high (with values higher than $0.8$ in many cases). The standard deviation is always low, indicating that the modularity of most of the instances in each set is close to its average. There are, however, some extreme cases for which the modularity is low; this can be observed in the minimum value of each set. Nevertheless, these extreme cases represent a very small fraction of each set, since the 10th percentile have a much higher value in most of the cases. In fact, the value of the 10th percentile is high enough to conclude that most industrial instances exhibit a clear community structure. Finally, although the computation of the community structure returned a \emph{memory-out} in some instances, it can be seen that this only happens in very few formulas.

%%%%%%%%%%%%%%%%%%%%%%%%%%%%%%%%%%%%%%%%%%%%%%%%%%%%%%%%%%%%%%%%%%%%%%%%%%%
\section{The Community Structure during SAT Solver Search}\label{sec-cdcl}
%%%%%%%%%%%%%%%%%%%%%%%%%%%%%%%%%%%%%%%%%%%%%%%%%%%%%%%%%%%%%%%%%%%%%%%%%%%

We want to investigate how CDCL techniques affect the community
structure of the formula. The natural question is: even if the
original formula shows a clear community structure, could it be the
case that this structure is quickly destroyed during the search
process? In other words, the learning mechanism \emph{increases} the
original formula with new learned clauses. How do these new clauses
affect the community structure of the formula?  Finally, even if the
value of the modularity is not altered, can it be the case that the
original partition of the formula is changed? In this section, we
investigate these phenomena.

\begin{table}[t]
\center
\begin{tabular}[b]{|rr||r|r|}
\hline
\multicolumn{1}{|c}{$n$} & $m/n$ & $Q_{orig}$ & $Q_{learned}$ \\
\hline
\hline
300 & 1.00      & \bf 0.459 & \bf 0.453 \\
300 & 2.00      & 0.291 & 0.291 \\
300 & 4.00      & 0.190 & 0.073 \\
300 & 4.25      & 0.183 & 0.041 \\
300 & 4.50      & 0.177 & 0.045 \\
300 & 6.00      & 0.150 & 0.120 \\
300 & 10.00    & 0.112 & 0.171 \\
\hline
\end{tabular}
\caption{Modularity $Q$ of random 3-CNF formulas with 300 variables
  varying the clause/variable ratio $m/n$, for original formulas
  ($Q_{orig}$), and formulas after adding all learned clauses kept by
  the solver when it finishes the search ($Q_{learned}$).}
\label{tab-dynamic-community-random}
\end{table}

Again, we start our analysis with random formulas.  In
Table~\ref{tab-dynamic-community-random}, we compare the modularity of
the original formula $Q_{orig}$ to the modularity of this formulas
augmented with all learned clauses that the solver is keeping when it
finishes the search $Q_{learned}$. The solver used to produce these
learned clauses is MiniSat~\cite{minisat}.  It is interesting to
observe that the closer to the peak transition region $m/n=4.25$, the lower
the modularity is with respect to the addition of learned clauses. A
possible explanation is that at the peak region we find the hardest
instances, and the harder an instance is, the more clauses connecting distinct
communities have to be learned, thus the lower the modularity becomes. Even
though, the modularity in all cases is very low, and the presence of
learned clauses does not contribute to increase the modularity of the
original formula (as expected for random instances).

\begin{table}[t]
  \begin{center}
    \tabcolsep 1.5mm
\begin{tabular}{|l||r|r|r|r|r||r|r|r|r|r|}\hline
&
\multicolumn{5}{c||}{VIG}&
\multicolumn{5}{c|}{CVIG}\\
Family  & $Q_{orig}$ & $Q_{prep}$ & $Q_{10^3}$ & $Q_{10^4}$ & $Q_{10^5}$ &
$Q_{orig}$ & $Q_{prep}$ & $Q_{10^3}$ & $Q_{10^4}$ & $Q_{10^5}$ \\\hline\hline
2d-strip-packing & \bf 0.942 & \bf 0.942 & \bf 0.942 & \bf 0.932 & \bf 0.884 & \bf 0.932 & \bf 0.928 & \bf 0.930 & \bf 0.926 & \bf 0.895 \\
bio & \bf 0.607 & \bf 0.549 & \bf 0.621 & \bf 0.619 & \bf 0.590 & 0.370 & 0.361 & 0.372 & 0.370 & 0.333 \\
crypto-aes & \bf 0.804 & \bf 0.752 & \bf 0.777 & \bf 0.737 & \bf 0.627 & \bf 0.610 & \bf 0.563 & \bf 0.598 & \bf 0.594 & \bf 0.552 \\
crypto-des & \bf 0.952 & \bf 0.929 & \bf 0.945 & \bf 0.929 & \bf 0.717 & \bf 0.498 & \bf 0.473 & \bf 0.503 & \bf 0.532 & \bf 0.496 \\
crypto-gos & \bf 0.639 & \bf 0.641 & \bf 0.621 & \bf 0.522 & \bf 0.424 & \bf 0.633 & \bf 0.623 & \bf 0.613 & \bf 0.531 & \bf 0.419 \\
crypto-md5 & \bf 0.784 & \bf 0.780 & \bf 0.850 & \bf 0.847 & \bf 0.825 & \bf 0.510 & \bf 0.544 & \bf 0.531 & \bf 0.538 & \bf 0.558 \\
crypto-sha & \bf 0.558 & \bf 0.641 & \bf 0.644 & \bf 0.641 & \bf 0.577 & \bf 0.562 & \bf 0.584 & \bf 0.584 & \bf 0.568 & \bf 0.475 \\
crypto-vmpc &  0.239 & 0.239 & 0.238 & 0.227 & 0.178 & 0.398 & 0.398 & 0.397 & 0.397 & 0.241 \\
diagnosis & \bf 0.932 & \bf 0.927 & \bf 0.932 & \bf 0.926 & \bf 0.871 & \bf 0.483 & \bf 0.444 & \bf 0.476 & \bf 0.478 & \bf 0.485 \\
hardware-bmc & \bf 0.922 & \bf 0.956 & \bf 0.923 & \bf 0.920 & \bf 0.835 & \bf 0.496 & \bf 0.468 & \bf 0.502 & \bf 0.496 & \bf 0.548 \\
hardware-bmc-ibm & \bf 0.971 & \bf 0.886 & \bf 0.970 & \bf 0.969 & \bf 0.962 & \bf 0.499 & \bf 0.432 & \bf 0.502 & \bf 0.501 & \bf 0.506 \\
hardware-cec  & \bf 0.857 & \bf 0.785 & \bf 0.853 & \bf 0.825 & \bf 0.765 & \bf 0.478 & \bf 0.461 & \bf 0.482 & \bf 0.476 & \bf 0.506 \\
hardware-velev & \bf 0.679 & \bf 0.678 & \bf 0.678 & \bf 0.677 & \bf 0.676 & \bf 0.486 & \bf 0.488 & \bf 0.484 & \bf 0.484 & \bf 0.490 \\
planning & \bf 0.865 & \bf 0.850 & \bf 0.856 & \bf 0.853 & \bf 0.834 & \bf 0.497 & \bf 0.496 & \bf 0.499 & \bf 0.499 & \bf 0.501 \\
scheduling & \bf 0.894 & \bf 0.781 & \bf 0.896 & \bf 0.885 & \bf 0.817 & \bf 0.474 & \bf 0.359 & \bf 0.454 & \bf 0.452 & \bf 0.487 \\
scheduling-pesp & \bf 0.780 & \bf 0.892 & \bf 0.780 & \bf 0.772 & \bf 0.662 & 0.359 & \bf 0.456 & 0.359 & \bf 0.431 & \bf 0.443 \\
software-bit-verif & \bf 0.878 & \bf 0.801 & \bf 0.872 & \bf 0.845 & \bf 0.728 & \bf 0.506 & \bf 0.568 & \bf 0.504 & \bf 0.509 & \bf 0.484 \\
termination & \bf 0.775 & \bf 0.695 & \bf 0.764 & \bf 0.674 & \bf 0.619 & \bf 0.525 & \bf 0.525 & \bf 0.521 & \bf 0.494 & \bf 0.456 \\
\hline
\end{tabular}
\end{center}
\caption{Modularity $Q_X$ of the formulas after X conflicts for VIG and CVIG models.}
\label{tab-dynamic-community-families}
\end{table}

Then, we analyze the evolution of the community structure for the case
of industrial SAT instances. As solving all industrial benchmarks is a
costly task (notice that some formulas are not even solved in the
competitions by any solver), we generate some set of learned clauses
running the solver for a fixed number of conflicts and augmenting the
original instances with the learned clauses the solver is keeping at
that moment. In this experiment, we use MiniSat, and we stop the
solver after $10^3$, $10^4$ and $10^5$ conflicts\footnote{These
  numbers of conflicts are not related to the number of conflicts
  required to solve the formula, but they increase in one order of
  magnitude, so they can be useful to analyze the evolution of the
  search.}.

In Table~\ref{tab-dynamic-community-families}, we show the values of
the modularities $Q_{orig}$ and $Q_{prep}$ of the original and
preprocessed formulas, and the modularities $Q_X$ of the formulas
after $X=10^3,10^4,10^5$ conflicts, for both the VIG and the CVIG
models.  We remark that these modularities are obtained with the LM
algorithm on the \emph{augmented} instances (i.e., original instances
with learned clauses).

We can observe that the modularity weakly decreases as we add learned
clauses, but it is still meaningful.  Therefore, learning does not
completely destroy the organization of the formula into (weakly)
connected communities. This means that LM is able to find a partition
of the (new) formula such that most of the edges connect variables of
the same community.

\begin{table}[t]
\begin{center}
\begin{tabular}{|l||r|r|r|r|}\hline
& \multicolumn{4}{c|}{VIG} \\
Family & $Q_{prep}$ & $Q^{part}_{10^3}$ & $Q^{part}_{10^4}$ & $Q^{part}_{10^5}$  \\ \hline
2d-strip-packing  &  \bf 0.942  &  0.272  &  0.209  &  0.132  \\
bio&  \bf 0.549  &  0.026  &  0.028  &  0.029  \\
crypto-aes &  \bf 0.752  &  \bf 0.346  &  \bf 0.324 &  0.250  \\
crypto-des  &  \bf 0.929  &  \bf 0.361 &  \bf 0.351  &  0.245 \\
crypto-gos  &  \bf 0.641  &  0.122  &  0.097  &  0.059  \\
crypto-md5  &  \bf 0.780  &  0.277  &  0.272  &  0.250  \\
crypto-sha  &  \bf 0.641  &  0.121  &  0.122  &  0.107 \\
crypto-vmpc  &  0.239  &  0.076  &  0.057  &  0.046  \\
diagnosis  &  \bf 0.927  &  0.308  &  0.327  &  0.306 \\
hardware-bmc &  \bf 0.886  &  \bf 0.715 &  \bf 0.702  &  \bf 0.632 \\
hardware-bmc-ibm  &  \bf 0.956  &  \bf 0.661  &  \bf 0.635 &  \bf 0.630  \\
hardware-cec  &  \bf 0.785  &  \bf 0.469   &  \bf 0.440  &  \bf 0.407 \\
hardware-velev  &  \bf 0.678  &  0.328   &  0.326 &  0.319  \\
planning  &  \bf 0.850  &  \bf 0.535  &  \bf 0.534  &  \bf 0.423  \\
scheduling  &  \bf 0.892  &  \bf 0.758  &  \bf 0.746  &  \bf 0.665  \\
scheduling-pesp  &  \bf 0.781  &  \bf 0.755  &  \bf 0.748  &  \bf 0.626 \\
software-bit-verif  &  \bf 0.801  &  \bf 0.569 &  \bf 0.547  &  \bf 0.449  \\
termination  &  \bf 0.695  &  \bf 0.428   &  0.372  &  0.313   \\ \hline
\end{tabular}
\end{center}
\caption{Modularity $Q^{part}_X$ of the formulas after X conflicts
  (for VIG), and using the partition of the original formula.}
\label{tab-dynamic-modmodules-families}
\end{table}

The question now is, even if the modularity does not decreases very
much, could it be the case that the communities have changed?  In
other words, can it be the case that there is still a clear community
structure but the partition of the formula into communities has
totally changed?

If a considerable part of learning is performed locally inside each
community, then the communities will not change. In VIG model, the set
of vertexes is always the same (even with the addition of learned
clauses). Notice that in this model, vertexes represent only
variables, so no learned clause creates new nodes. However, these
learned clauses do create new edges between the existing
nodes. Therefore, we can use modularity as a \emph{quality measure} to
see how \emph{internal} a learned clause is. Notice that modularity is
a function of two parameters: a graph, and a partition of it. For a
given partition of a graph, a new edge will increase the modularity
iff it connects two nodes of the same community, otherwise modularity
will decrease. Thus, using the partition of the original formulas, we
can see if learning acts \emph{internally} (i.e., connecting variables
of the same community), or if it tends to connect variables of
different communities.

We have conducted another experiment to see how learning changes such
partition. In this experiment, we use the same formulas than before (original formulas
augmented with learned clauses kept by the solver after $10^3$, $10^4$
and $10^5$ conflicts),
but in contrast to the previous experiment, we use now the partition of the original formulas
(i.e., without learned clauses).
We refer this modularity as $Q^{part}$.
Notice that
in the case we do not run the LM algorithm to compute a (possibly) new
partition, but we give explicitly that partition (i.e., the partition of the original formula). Moreover, we can
only use the VIG since the set of nodes is the same in both formulas:
original instances and formula augmented with learned clauses. We recall that using the CVIG, each new
(learned) clause adds a new clause-node to the graph.

In
Table~\ref{tab-dynamic-modmodules-families}, we show the result of the
modularity $Q^{part}$.  The analysis of this experiment shows us that
there is a drop-off in the modularity as we incorporate more learned
clauses. In other words, the partition of the formula is changing.
This means that, if we used explicitly the community structure to
improve the efficiency of a SAT solver, to overcome this problem we
would have to recompute such a partition (after some number of conflicts)
to adjust it to the modified formula.

Let us represent this effect using the graph of
communities\footnote{We cannot directly represent the VIG due to its
  large number of nodes (variables).}. This graph is built as
follows. All nodes of the VIG (variables) that belong to the same
community are merged into a single node in the graph of communities,
and weighted edges are updated accordingly. The weight of the edge
connecting communities $A$ and $B$ is the addition of the weights of
the edges connecting one node from community $A$ and one node from community $B$ in the original graph.

\begin{figure}[t]
\hbox to \columnwidth {
\mbox{}\hss
\includegraphics[width=0.33\textwidth]{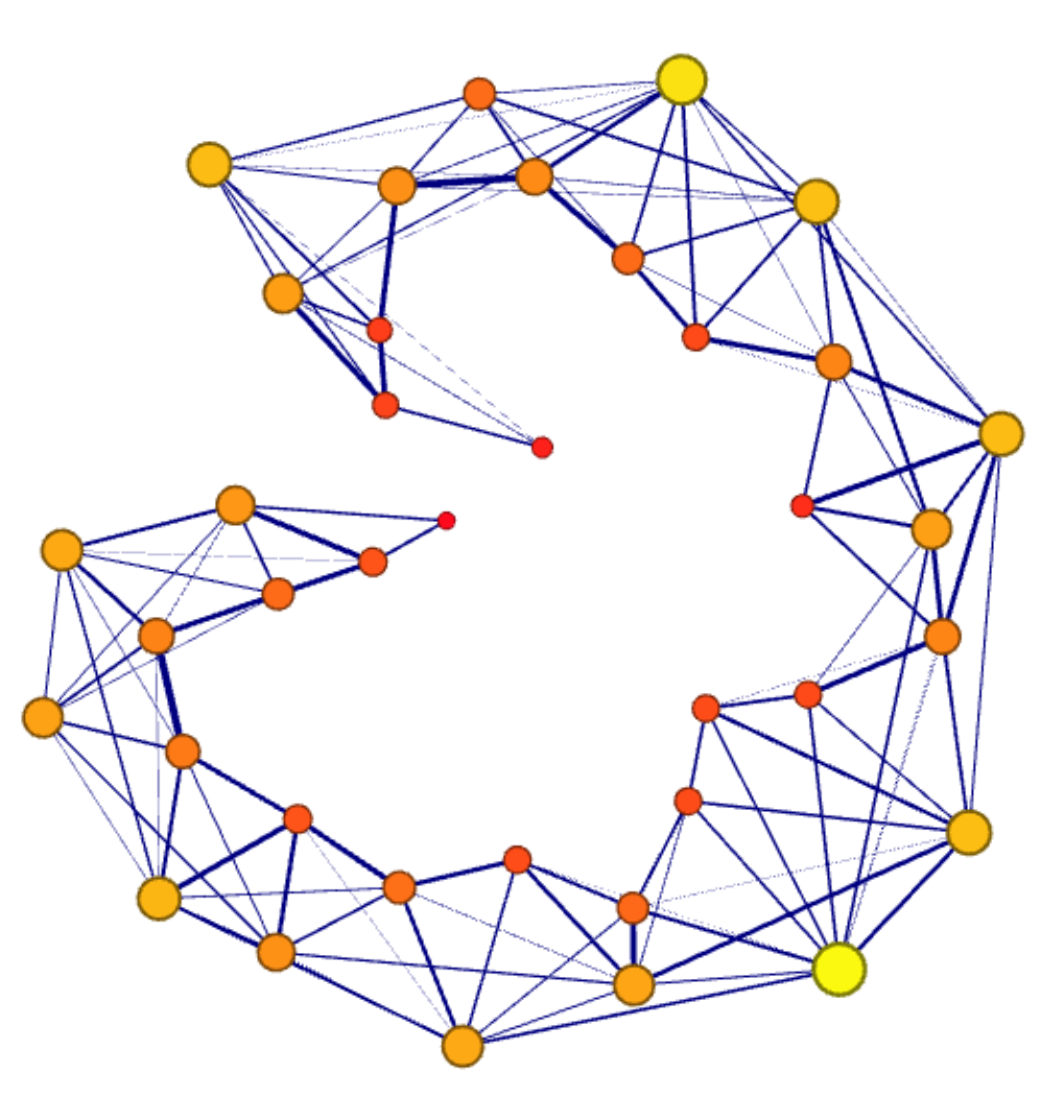}
\hss
\includegraphics[width=0.33\textwidth]{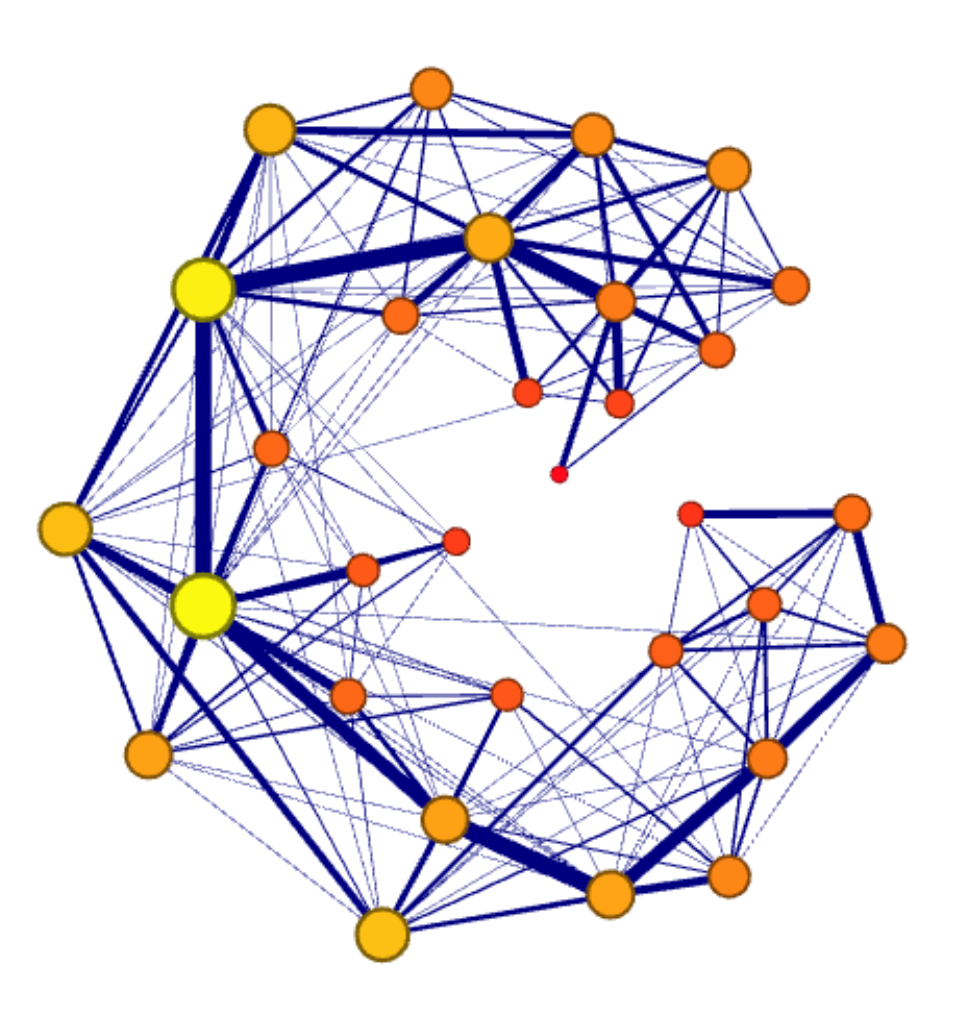}
\hss
\includegraphics[width=0.33\textwidth]{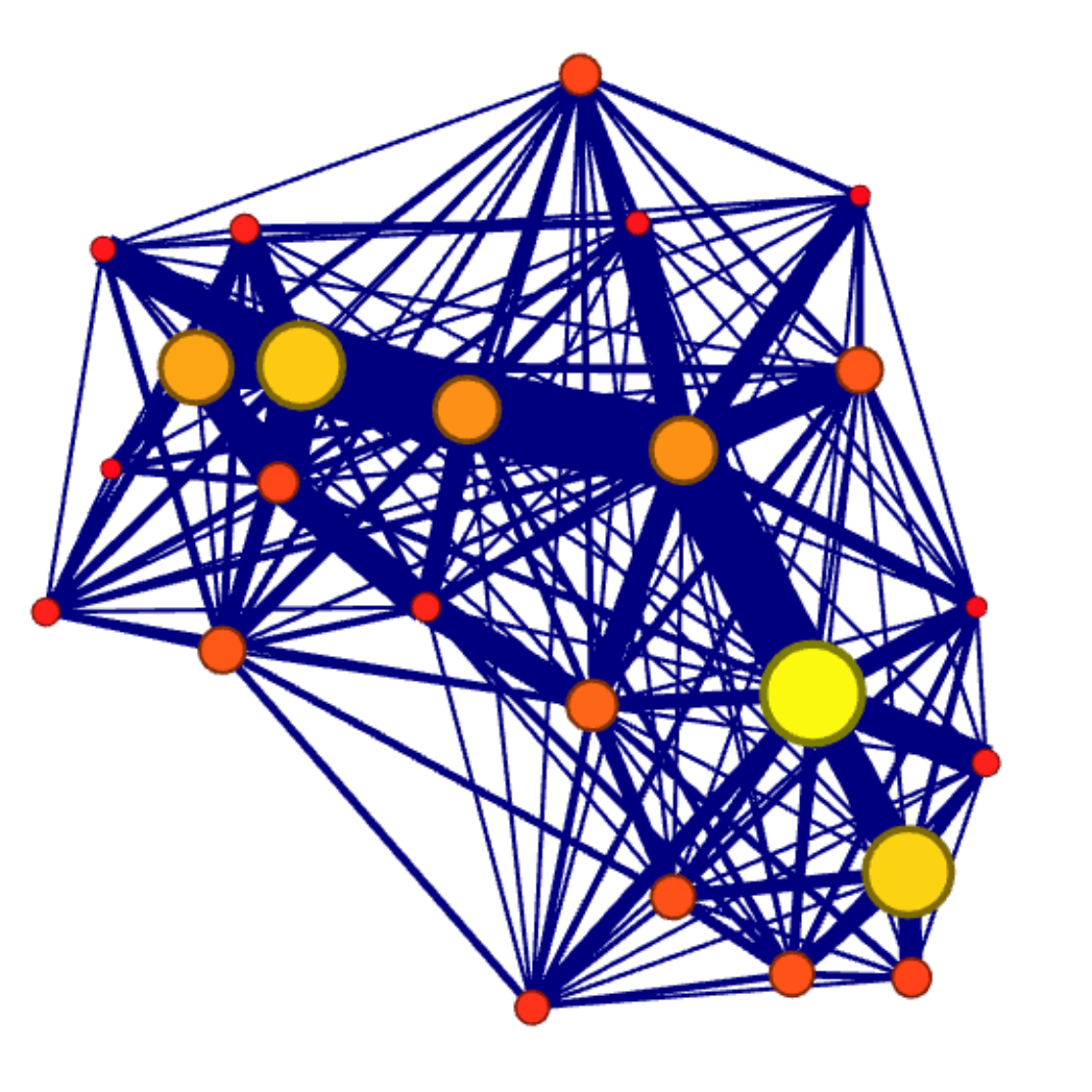}
\hss\mbox{}
}
\caption{Graph of communities of the instance {\tt ibm-2002-22r-k60}:
  original formula (left), solved formula considering \emph{small}
  learned clauses (center), and solved formula considering \emph{small}
  and \emph{medium-sized} learned clauses (right). Nodes and edges are
  accordingly scaled by community size and weight, respectively.}
\label{fig-example}
\end{figure}

In Figure~\ref{fig-example} (\emph{left}), we represent the graph of
communities of the industrial formula {\tt ibm-2002-22r-k60}. This
instance has a modularity $Q=0.91$ and $35$
communities. Glucose~\cite{Laurent09} solved this formula keeping a
total of $504964$ learned clauses. We can recompute the graph of
communities after adding some of these learned clauses to the original
instance. In Figure~\ref{fig-example} (\emph{center} and
\emph{right}), we represent the graph of communities after adding
\emph{small} learned clauses (up to 10 literals), and
\emph{medium-sized} learned clauses (up to 50 literals),
respectively.\footnote{As each clause of length $l$ generates $l
  \choose 2$ edges, it is hard to compute these graphs using
  \emph{long} clauses.} The modularity of these augmented instances is
respectively $0.87$ and $0.82$, and the number of communities $29$ and
$24$. In these graphs of communities, the node size is scaled
according to the number of variables that belong to each
community. Also, edges are scaled by their weights. Notice that edges
weights are computed using the weights of the VIG (i.e., taking into
account the length of the original clauses). The community structure is clear
in all of these three graphs. However, as we consider more learned
clauses, we can observe two phenomena. First, the number of
communities (number of nodes in the graph of communities)
decreases. This means that variables that originally belonged to
distinct communities are now grouped into the same community. Second,
the weight of the inter-communities edges increases. Therefore, from
the two previous effects, we observe that the solver prefers to learn
clauses containing variables of distinct (original) communities. For
these reasons, in general clause learning contributes to decrease the
modularity.

Finally, we want to determine how much each learned clause contributes
to destroy the original organization of the formula. To this purpose,
we can measure the increase of the modularity $\Delta Q$ that each
learned clause produces. Notice that $\Delta Q$ is positive when most
of the new edges generated by such clause connect nodes (variables) of
the same community. Otherwise, $\Delta Q$ is negative.

\begin{figure}[!t]
  \hbox to \columnwidth {
    \includegraphics[width=0.49\textwidth]{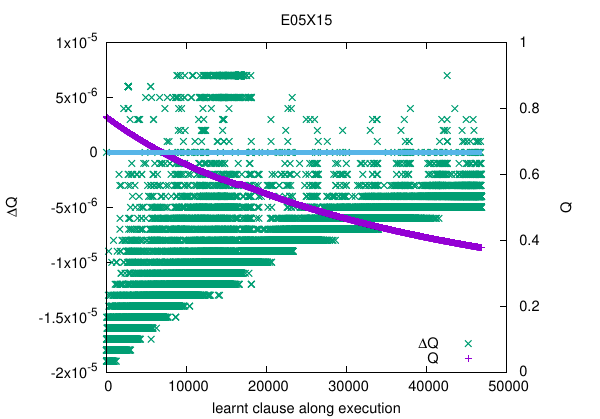}
    \includegraphics[width=0.49\textwidth]{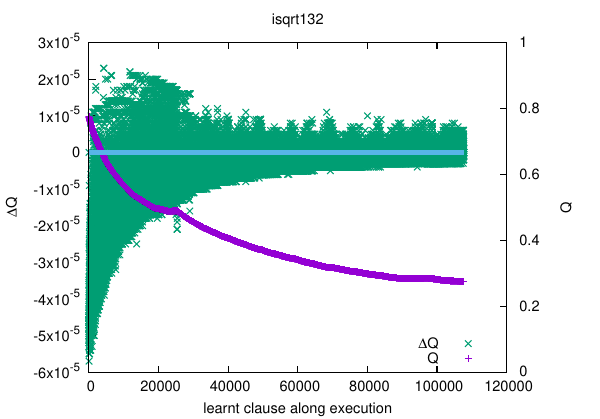}
  }
\caption{Impact of adding learned clauses on modularity, in instances
  {\tt E05X15} (left) and {\tt isqrt1\_32} (right). Each point
  $(x,y)$, with $y$ measured in the left $Y$ axis, represents a clause
  learned at instant $x$ and increasing $Q$ on $y$.  We also represent
  the evolution of the modularity $Q$ (using the right $Y$ axis).}
  \label{fig-example-2}
  \end{figure}
  
After an extensive experimentation on a subset of UNSAT industrial
instances, we see that, in general, each learned clause produces a
decrease of the modularity (i.e., $\Delta Q <0$), but this decrease is
very small (i.e., $\Delta Q \approx 0$).
  
In Figure~\ref{fig-example-2}, we represent this analysis for the
industrial instances {\tt E05X15} and {\tt isqrt1\_32}. Each point
$(x,y)$, with $y$ measured in the left $Y$ axis, represents a clause
learned at instant $x$ and increasing $Q$ on $y$.  We also represent
(using the right $Y$ axis) the current value of the modularity $Q$
using the original partition of variables, along the execution. We can
see that the contribution to increase or decrease the modularity is
very small (i.e., $\Delta Q \approx 0$).  Also, even when some learned
clauses contribute to increase the value of $Q$, most of them do not
(i.e., $\Delta Q < 0$), and thus $Q$ tends to decrease.

Although we only represent these two benchmarks, we emphasize that we observed similar results in most industrial SAT instances
studied.
Therefore, we can conclude that, in general, learned clauses contribute to
destroy the (original) community structure of the formula. It is not due to
some particular clauses but rather a general phenomenon of the learning
mechanism.

%%%%%%%%%%%%%%%%%%%%%%%%%%%%%%%%%%%%%%%%%%%%%%%%%%%%%%%%%%%%%%%%%%%%%%%%%%%
\section{A Modularity-based Preprocessor to Detect Relevant Learned Clauses}\label{sec-modprep}
%%%%%%%%%%%%%%%%%%%%%%%%%%%%%%%%%%%%%%%%%%%%%%%%%%%%%%%%%%%%%%%%%%%%%%%%%%%

Learning new clauses during the execution of the SAT solver is undoubtedly a crucial component of state-of-the-art CDCL SAT solvers; their success would not have been possibly achieved without the incorporation of clause learning within them.
In fact,~\citeA{SakallahM11} conducted an analysis to determine which CDCL component contributes in a higher degree to the success of these solvers, and concluded it is indeed clause learning.

Clause learning can be seen as a two-sided effect. On the one hand, adding new redundant\footnote{Notice that learned clauses are redundant by definition, hence not strictly necessary.} clauses avoids the solver to explore the same subspace(s) in the future. However, this makes the management of the clause database harder, since it can grow exponentially. This especially affects Unit Propagation (UP), since it is more costly to detect which clauses may propagate at each step. For this reason, memory was an important issue in the early versions of CDCL SAT solvers~\cite{grasp,MoskewiczMZZM01,minisat}, and some heuristics were proposed to remove some learned clauses periodically. On the other hand, these learned clauses guide the search to a particular subspace, where \emph{easier} proofs are hopefully found. With this aim,~\citeA{Laurent09} propose LBD --Literal Block Distance-- as a measure of quality of learned clauses, and they implement a very aggressive clause removal policy in their CDCL solver Glucose. Notice that in some cases, Glucose removes up to 95\% of the learned clauses.  Therefore, the objective is not any longer to keep as many learned clauses as possible (to achieve a good pruning of the search) without exceeding a good UP rate, but to keep as few --but relevant-- learned clauses as possible (hence a good UP rate is certainly preserved) and achieve a good pruning with them.

The LBD of a learned clause is the number of distinct decision levels of its variables. The idea behind LBD is that literals propagated at the same decision level are tightly connected, and they may often be propagated once and again together. An interesting case are the learned clauses with LBD 2 (called \emph{glue clauses}), which are kept forever in Glucose. Recently, it was shown that the LBD value is correlated to the number of distinct communities in the clause~\cite{NewshamGFAS14}. In this section, we show that we can use the community structure of the formula to detect relevant learned clauses.

\begin{algorithm}[t]
\label{alg-solver}
\KwIn{SAT Instance $\Gamma$}
\KwOut{SAT Instance $\Gamma'$}
$\Gamma' := \Gamma$\;
$C := communityStructure(\Gamma)$\;
\ForEach{pair $(c_i,c_j)$ of connected communities of $C$}{
	Solver $s$\;
	$s.solve(c_i \cup c_j)$\;
	\If{$s == UNSAT$}{
		\KwRet{$\varnothing$}\;
	}
	$\Gamma' := \Gamma' \cup s.learnedClauses$\;
}
\KwRet{$\Gamma'$}\;
\caption{Modularity-based SAT Instance Preprocessor ({\bf modprep})}
\end{algorithm}

\citeA{SAT15} perform an interesting experiment to evaluate the contribution of learned clauses to the success of the solver, distinguishing between satisfiable and unsatisfiable SAT formulas. In particular, they compute the runtime $t$ of a CDCL SAT solver on a set of benchmarks.\footnote{In fact, they measure the evolution of the solver in terms of number of generated conflicts, but the conclusion is essentially the same using runtime.} Then, they repeat the same execution of the solver, stopping at $p\,t$, for some fixed $p \in (0,1)$, and generating a new formula containing all original clauses plus all learned clauses stored in the solver at that stage. Then, they compute the time $t'$ needed to solve these augmented formulas. They detect that, for most unsatisfiable formulas, $t \approx p\,t + t'$. However, in the case of satisfiable formulas $t$ and $p\,t+t'$ are very different (in most cases $p\,t+t' \gg t$, especially for big values of $p$). This is, clause learning is not the only component in play. Notice that this experiment is equivalent to removing all activity counters used by the heuristic after time $p\,t$, and this may dramatically worsen the performance of the solver when the formula is satisfiable. So, the natural question is whether it is possible to detect very relevant clauses that also help the solver to guide the search in satisfiable instances.

In the previous sections, we concluded that real-world SAT instances usually have high modularity, and that clause learning tends to destroy their (original) community structure, but it does it slowly. This suggests that \emph{good} learned clauses must contribute to destroy this structure in a \emph{low} degree. This is, these good clauses are precisely those that connect few communities. In this section, we present an application that exploits the community structure to detect relevant learned clauses, and we show that detecting these clauses may result into an improvement on the performance of the SAT solver in satisfiable instances without altering its performance on unsatisfiable ones. We present this application as a preprocessing technique, so it can be easily incorporated into any existing solver.

This preprocessor, called \emph{Modularity-based SAT Instance Preprocessor} ({\bf modprep}), is presented in Algorithm~\ref{alg-solver}. It augments the original formula with some learned clauses based on its community structure. This algorithm proceeds as follows. First, it computes the community structure of the original
formula (line~2).\footnote{In particular, we compute the community structure of the VIG (which assigns each variable to a certain community), and then assign each clause to the most frequent community among its variables (in case of ties, the clause is randomly assigned to one of the most frequent communities among its variables). Compared to the CVIG, the obtained results with VIG are very similar, but it is more efficient to compute them in practice due to the smaller size of VIG (VIG has a smaller number of nodes and edges).}
This way, the set of clauses of the formula is split into disjoint communities.
Then, for each pair of connected communities,\footnote{Two communities $a$ and $b$ are connected if there exists at least one variable that appears in a clauses of community $a$ and in a clause of community $b$.} it creates a
subformula containing all the clauses belonging to both communities, and solves it (line~5). 
If this subformula is UNSAT, it returns the empty clause, i.e., the original formula is unsatisfiable. Otherwise, the original instance is augmented with the clauses the solver learned for solving such subformula (line~8). Finally, it returns this augmented instance.

In our experiments, the core SAT solver used by our preprocessing technique (used to solve all subformulas --see Line~4 in Alg.~\ref{alg-solver}--) is MiniSat. A natural question is whether the number of clauses learned in this process depends on the solver used by the preprocessor. In order to check it, if we use Glucose instead of MiniSat for solving all subformulas, the resulting number of learned clauses is very similar, with no significant difference between them. This is because all subformulas are extremely easy, and thus, the choice about the solver used in Alg.~\ref{alg-solver} does not seem to alter the output of the algorithm in terms of number of learned clauses computed.

Notice that the previous algorithm imposes a very strong condition,
which is solving \emph{all} subformulas between two connected
communities and keeping \emph{all} learned clauses found in this
process. This could be further refined. For example, in case we incorporated the number of communities in the clause as an heuristics for deciding clause deletion, instead of using it in the preprocessing, we conjecture that improvements would be even greater than what we observe in this experiment. Moreover, this preprocessing step
could be heuristically applied during the search in the flavor of
inprocessing approaches~\cite{inprocessing}. 

Although we will show that this approach works
experimentally, we may wonder why these learned clauses indeed improve
the performance of the solver. It is  worth noticing that, by construction, these learned clauses are composed of at most 2 communities, and thus are clearly related to the notion of \emph{glue clauses} aforementioned. In addition, as shown previously, learned clauses contribute to destroy
the original community structure, but do it slowly.
We consider that the natural case to achieve such a ``\emph{slow destruction}'' behavior is to learn clauses connecting pairs of communities.
Notice that
a solver not aware of the community structure may remove them, unless,
as we do, these clauses are added in a preprocessing step as original
clauses, so the solver is forced to keep them.

%%%%%%%%%%%%%%%%%%%%%%%%%%%%%%%%%%%%%%%%%%%%%%%%%%%%%%%%%%%%%%%%%%

\subsection{Experimental Evaluation}

We now present an experimental evaluation of the
modularity-based preprocessor \emph{modprep}.
Since this tool can be easily incorporated into any SAT solver, in our evaluation we select a number of CDCL solvers and evaluate their performance with and without using \emph{modprep}. In particular, we compare the running time of solving an instance by a certain solver (without using \emph{modprep}) with respect to the running time of this preprocessor on that instance plus the solving time of that solver on the output (augmented) instance returned by \emph{modprep}.

When \emph{modprep} fails to compute the augmented formula $\Gamma'$ (see Alg.~\ref{alg-solver}, line~9), it returns the input formula $\Gamma$. This happens, for instance, when the formula is so huge that the systems does not have enough memory to compute its community structure (i.e., \emph{memory-out}). Therefore, in this case, both solvers (i.e., with and without \emph{modprep}) are solving the same formula $\Gamma$, but their running times may differ, since we take into account the running time spent by the preprocessor.

\begin{table}[t]
\center
\begin{tabular}{|l||r|r||r|r||r|r||r|r|}
\hline
 & \multicolumn{2}{c||}{Glucose} &  \multicolumn{2}{c||}{MapleSAT} &  \multicolumn{2}{c||}{Lingeling} &  \multicolumn{2}{c|}{MiniSat}\\
\hline
\emph{modprep} used? & \multicolumn{1}{c|}{No} & \multicolumn{1}{c||}{Yes} & \multicolumn{1}{c|}{No} & \multicolumn{1}{c||}{Yes} & \multicolumn{1}{c|}{No} & \multicolumn{1}{c||}{Yes} & \multicolumn{1}{c|}{No} & \multicolumn{1}{c|}{Yes} \\
\hline

& \multicolumn{8}{c|}{Only SAT} \\
\hline
SAT Race 2010 		& \underline{\bf 23} & 22 	& 22 & 22			 				& 21 & 21			 		& 22 & 22 \\
SAT Competition 2011 	& \bf 95 & 91 			& 103 & \bf \underline{105} 			& \bf 96 & 88 				& 94 & \bf 98 \\
SAT Challenge 2012 	& 248 & 248	 		& 256 & \bf \underline{258} 			& \bf 245 & 238 			& 248 & \bf 251 \\
SAT Competition 2013 	& \bf 129 & 124 		& 124 & \bf \underline{131} 			& \bf 103 & 97 				& 116 & \bf 125 \\
SAT Competition 2014 	& 104 & \bf 110 		& 111 & \bf \underline{116} 			& \bf 103 & 102 			& 97 & \bf 99 \\
SAT Race 2015 		& 138 & \bf 140 		& 155 & \bf \underline{157} 			& \bf 143 & 139 			& 130 & \bf 135 \\
SAT Competition 2016 	& 62 & \bf 64 			& \underline{67} & \underline{67}		& \bf 58 & 53 				& \bf 63 & 60 \\
SAT Competition 2017 	& 79 & \bf 80 			& \underline{\bf 97} & 96  				& \bf 79 & 70 				& 88 & \bf 89 \\
\hline
Total SAT 				& 878 & \bf 879 		& 935 & \bf \underline{952} 			& \bf 848 & 808 			& 858 & \bf 879 \\
\hline

& \multicolumn{8}{c|}{Only UNSAT} \\
\hline
SAT Race 2010 		& \bf 63 & 62 			& \underline{64} & \underline{64} 		& 62 & 62			 			& 58 & 58 \\
SAT Competition 2011 	& \bf 123 & 120 		& \bf 125 & 123 					& \underline{\bf 131} & 130 		& \bf 98 & 96 \\
SAT Challenge 2012  	& 304 & 304	 		& 312 & 312	 					& 317 & \bf \underline{318} 		& 255 & \bf 259 \\
SAT Competition 2013 	& \bf 121 & 115 		& 123 & \bf \underline{124} 			& \bf 106 & 105 				& 67 & \bf 69 \\
SAT Competition 2014 	& 122 & 122	 		& \bf 123 & 120 					& \underline{135} & \underline{135} 	& \bf 71 & 70 \\
SAT Race 2015 		& 104 & \bf 105 		& \bf 107 & 106			 			& \underline{109} & \underline{109}	& 73 & \bf 75 \\
SAT Competition 2016 	& \bf 86 & 81 			& \bf 82 & 81			 			& 95 & \bf \underline{97} 			& 61 & 61 \\
SAT Competition 2017 	& 91 & 91	 			& \underline{94} & \underline{94}		& \underline{94} & \underline{94}	& 68 & \bf 70 \\
\hline
Total UNSAT 			& \bf 1014 & 1000 		& \bf 1030 & 1024 					& 1049 & \bf \underline{1050} 		& 751 & \bf 758 \\
\hline

& \multicolumn{8}{c|}{SAT + UNSAT} \\
\hline
SAT Race 2010 		& \underline{\bf 86} & 84 	& \underline{86} & \underline{86} 		& 83 & 83			 		& 80 & 80 \\
SAT Competition 2011 	& \bf 218 & 211 		& \underline{228} & \underline{228}		& \bf 227 & 218 			& 192 & \bf 194 \\
SAT Challenge 2012 	& 552 & 552	 		& 568 & \bf \underline{570} 			& \bf 562 & 556 			& 503 & \bf 510 \\
SAT Competition 2013 	& \bf 250 & 239 		& 247 & \bf \underline{255} 			& \bf 209 & 202 			& 183 & \bf 194 \\
SAT Competition 2014 	& 226 & \bf 232 		& 234 & \bf 236 					& \bf \underline{238} & 237 	& 168 & \bf 169 \\
SAT Race 2015 		& 242 & \bf 245 		& 262 & \bf \underline{263} 			& \bf 252 & 248 			& 203 & \bf 210 \\
SAT Competition 2016 	& \bf 148 & 145 		& \bf 149 & 148						& \underline{\bf 153} & 150 		& \bf 124 & 121 \\
SAT Competition 2017 	& 170 & \bf 171 		& \bf \underline{191} & 190 			& \bf 173 & 164 			& 156 & \bf 159 \\
\hline
Total SAT+UNSAT 		& \bf 1892 & 1879 		& 1965 & \bf \underline{1976} 			& \bf 1897 & 1858 			& 1609 & \bf 1637 \\
\hline

\end{tabular}
\caption{Number of solved instances with and without the modularity-based preprocessor \emph{modprep}, on the application instances of all SAT Competitions from 2010 to 2017. For each solver and competition, in bold it is marked the best choice among using or not using \emph{modprep}. The best solver for each competition (row) is underlined.}
\label{tab-results-all}
\end{table}

In our experimental evaluation, we use all application SAT benchmarks used in the SAT Competition from 2010 to 2017 (both included). Recall that this set contains a total of 2550 SAT instances. We evaluate four well-known CDCL SAT solvers. Namely, they are Glucose~\cite{Laurent09}, MapleSAT, using its version with LRB~\cite{LiangGPC16}, Lingeling~\cite{lingeling}, and MiniSat~\cite{minisat}. Notice that Glucose, MapleSAT and Lingeling have been ranked as some of the best solvers in the last competitions, whereas MiniSat is possibly the most famous CDCL SAT solver, in which many other solvers are based on. The experiments were carried out limiting the memory usage to 16GB, and using a timeout of 5000~seconds. The preprocessor \emph{modprep} is executed with a timeout of 100 seconds

First, we evaluate how expensive is running the preprocessor \emph{modprep} described in Alg.~\ref{alg-solver}. Notice that this algorithm can be split into two steps: i)~computing the community structure to partition the input formula into subformulas; and ii)~solving them. On the set of 2550 application SAT instances we use in our experiments, this tool is able to correctly compute the community structure of 2247 formulas. This represents a 88.15\% of the set. The average runtime is 7.08 seconds, with a median of 1.37 seconds and a percentile 95 of 40.56 seconds. Therefore, this is a very fast step. The second step is a bit slower, so it times out in some instances. In particular, this tool is able to finish in 2158 instances (a 84.66\% of the total) within the timeout of 100 seconds. In this case, the average runtime is 12.36 seconds, with a median of 3.17 seconds and a percentile 95 of 65.22 seconds. Therefore, for most of the instances, running \emph{modprep} is fast.

As said before, the preprocessor \emph{modprep} sometimes times out. We have detected that in some (but very few) cases, a certain subformula might be extremely hard, so solving such a subformula is almost as hard as solving the original formula. Notice that if the preprocessor fails to finish, it returns the original instance. However, the running time of a certain solver using \emph{modprep} may differ of the one without using it, although they may be solving the same instance. This is due to the time spent by the preprocessor.

In Table~\ref{tab-results-all}, we represent the results of this  experiment. In particular, we detail the number of solved instances by each of the four solvers (Glucose, MapleSAT, Lingeling and MiniSat) with and without using the modularity-based preprocessor \emph{modprep}, distinguishing between three categories: SAT+UNSAT, SAT only and UNSAT only, on application benchmarks used in the SAT Competitions from 2010 and 2017.

\begin{figure}[t!]
\includegraphics[width=0.9\textwidth]{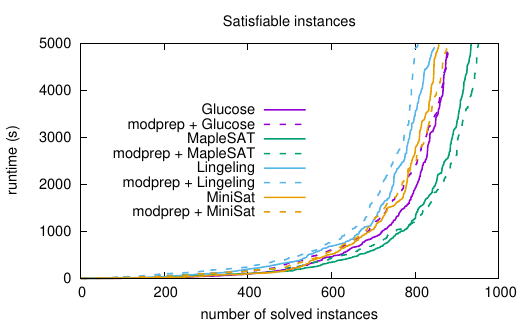}
\includegraphics[width=0.9\textwidth]{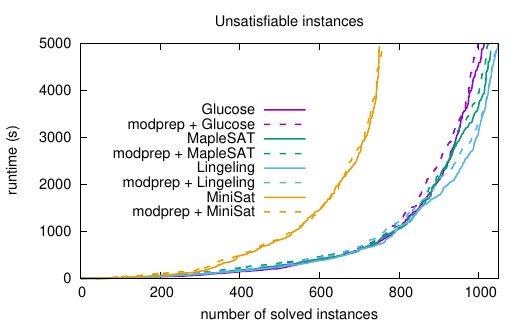}
\caption{Cactus plot representing the results of the solvers Glucose, MapleSAT, Lingeling and MiniSat with and without using the modularity-based preprocessor \emph{modprep}, on the aggregated set composed of all application instances from the SAT Competitions from 2010 to 2017, distinguishing between satisfiable (top) and unsatisfiable (bottom) instances.}
\label{fig-cactus-all}
\end{figure}

For satisfiable instances, we achieved in many cases an improvement on the performance of Glucose, MapleSAT and MiniSat. For instance, Glucose improves its performance on 4 of these 8 competitions sets (and ties in another one). In fact, the best choice among the 8 evaluated solvers (4 solvers with and without \emph{modprep}) is, in most of the cases, MapleSAT enhanced with our technique.

In the case of unsatisfiable instances, we mostly achieved improvements in MiniSat, and in some cases, in the other three solvers. However, the differences between using and not using the preprocessor seem to be small. In this case, the best solver in the aggregated set of all competitions is Lingeling enhanced with \emph{modprep}.

On the set of all instances (i.e., SAT+UNSAT), we observe that, in general, the solvers MapleSAT and MiniSat, and in some cases the solver Glucose, improve their performance. On the contrary, we do not achieve better results for Lingeling, because the performance on satisfiable instances is much worse when the solver is enhanced with the preprocessor, so the good results on unsatisfiable instances are not enough to achieve an overall improvement in this solver. However, in the other three solvers, we can observe clear improvements in many cases, for the union of satisfiable and unsatisfiable instances. In most of the cases, this is due to the important improvements achieved on satisfiable instances.

In Fig.~\ref{fig-cactus-all}, we represent the cactus plot (i.e., number of instances solved in a certain wall clock time) of these four solvers  with and without using \emph{modprep}, on the aggregated set of all SAT Competitions between 2010 and 2017, distinguishing between satisfiable and unsatisfiable instances. We recall that the running times of the solvers using \emph{modprep} include the running time of the preprocessor. 

On unsatisfiable instances, incorporating the preprocessor does not seem to affect solver performance. This observation suggests that detecting relevant learned clauses may be especially important for satisfiable formulas, as~\citeA{SAT15} suggested.

On satisfiable instances, however, it can be observed clear improvements in MapleSAT and MiniSat when the preprocessor is incorporated, and a very small improvement in Glucose. But this figure allows us to see another interesting observation: there is a tendency of higher improvements for larger timeouts.  This is possibly due to the time spent by the preprocessor in very easy instances. For those formulas, the preprocessor only adds an extra overhead to the solving time. However, it seems that the improvements of using the preprocessor become more clear in harder instances, i.e., the ones that require longer runtimes to be solved. This is clearly seen for MapleSAT and MiniSat, and it seems to be the case for Glucose. In fact, preliminary results with a much longer timeout confirm this observation for the solver Glucose as well~\cite{SAT15}.

These results suggest that the community structure of industrial SAT formulas is not a simple artifact, but it captures a relevant feature of the underlying structure of these instances, which partially explains the distinct performance of SAT solvers on random and industrial formulas, and which can be exploited by modern SAT solvers.

%%%%%%%%%%%%%%%%%%%%%%%%%%%%%%%%%%%%%%%%%%%%%%%%%%%%%%%%%%%%%%%%%%%%%%%%%%%
\section{Conclusions}\label{sec-conclusions}
%%%%%%%%%%%%%%%%%%%%%%%%%%%%%%%%%%%%%%%%%%%%%%%%%%%%%%%%%%%%%%%%%%%%%%%%%%%

Inspired by \emph{complex networks}, we have studied one decisive
feature of the \emph{underlying structure} of industrial SAT formulas,
representing them as graphs. The classical Erd\"os-R\'enyi model for
generating random graphs cannot be used for studying \emph{real-world}
networks, since they exhibit some particular \emph{structural
  properties}. In the case of SAT instances, this
model is appropriate to study random formulas, but not for modeling
industrial instances. These instances are characterized by a
particular structure, which may explain their distinct nature
w.r.t. random formulas. In particular, we have analyzed the
\emph{community structure}, or the \emph{modularity}, of these
benchmarks. Moreover, we have studied how this structure evolves during the
execution of a CDCL SAT solver. Finally, inspired by the observations on our analysis, we have proposed
an application that explicitly exploits the community structure of the formula to detect relevant clauses, and
learning those clauses results into an overall improvement on the performance of several CDCL SAT solvers, especially on satisfiable formulas.

We have seen that most industrial instances exhibit a clear community
structure (whereas random formulas do not). This means that we can
find a partition of the formula into communities in which variables
are highly interconnected. In general, industrial formulas have a
exceptionally high modularity, greater than $0.8$ in many
cases. Notice that in other kind of networks, values greater than
$0.7$ are rare.

Also, we have analyzed on this
structure the effect of learning new clauses during the search of the SAT solver.
Interestingly, most of the learned clauses tend to connect
variables of different communities. As a consequence, learning new
clauses destroys the original structure of the formula. However, this
occurs very slowly, since each learned clause contributes very little
to the decrease of the modularity. This behavior is observed in all
benchmarks analyzed. Therefore, it seems that the solver performs the
search destroying the original community organization of the formula.

Finally, we have presented a preprocessing technique that modifies any input formula
by adding to it some relevant learned clauses, found by exploiting the community structure
of the instance. In particular, we use the community structure to split the 
formula into many subformulas, each containing a pair of connected communities. These subformulas are
solved in order to learn the set of clauses that is added to the original instance.
Notice that these clauses only contain, at most, variables of two distinct communities, and hence are closely related to the concept of \emph{glue clauses} used in the CDCL SAT solver Glucose (i.e., learned clauses whose LBD value is 2). In practice, this preprocessing step is efficiently computed. Based on the results of our empirical evaluation, we conclude that enhancing a SAT solver with such a preprocessing techniques is beneficial in many cases, especially for satisfiable instances.

We think that the present study provides a step towards an
explanation of why some SAT solvers perform better on industrial
instances, and others on random SAT formulas. Moreover, the better
understanding of this structure in real-world instances has led to the
improvement of existing SAT
solvers~\cite{DMartinsML13,NevesMJLM15,SonobeKI14}.
This analysis also serves as basis for new random SAT generation
models that produce more realistic pseudo-industrial random
instances~\cite{IJCAI15,AIJ4,IJCAI17}. This problem is distinguished as one of the 10 challenge
problems in
SAT~\cite{tenchallenges1,tenchallenges2,tenchallenges3,tenchallenges4}.
Understanding the structure of industrial instances is a first step
towards the development of random instance generators, reproducing the
features of industrial instances. These generators can be used to
support the testing of industrial SAT solvers under development.

\acks{
This work is partially supported by the EU H2020 Research and Innovation Programme under the LOGISTAR project (Grant Agreement No. 769142), MINECO-FEDER projects RASO (TIN2015-71799-C2-1-P/2-P) and TASSAT3 (TIN2016-76573-C2-2-P), the Spanish Ministerio de Econom\'ia y Competitividad under the EXASOCO project (ref. PGC2018-101216-B-I00), including European Regional Development Funds (ERDF). The third author is also supported by a MICINN Juan de la Cierva fellowship (ref. FJCI-2017-32420).
}

%\section*{References}
\bibliographystyle{theapa}
\bibliography{jair}

\end{document}